\documentclass[journal]{IEEEtran}
\usepackage{amsmath,amsfonts}
\usepackage{algorithmic}
\usepackage{algorithm}
\usepackage{array}
\usepackage[caption=false,font=normalsize,labelfont=sf,textfont=sf]{subfig}
\usepackage{textcomp}
\usepackage{stfloats}
\usepackage{url}
\usepackage{verbatim}
\usepackage{graphicx}
\usepackage{cite}
\usepackage[T1]{fontenc}
\usepackage{multirow, makecell}
\usepackage{color,xcolor}
\usepackage{booktabs}
\usepackage{balance}

\begin{document}

%\title{A Sample Article Using IEEEtran.cls\\ for IEEE Journals and Transactions}

%\author{IEEE Publication Technology,~\IEEEmembership{Staff,~IEEE,}
        % <-this % stops a space
%\thanks{This paper was produced by the IEEE Publication Technology Group. They are in Piscataway, NJ.}% <-this % stops a space
%\thanks{Manuscript received April 19, 2021; revised August 16, 2021.}}

% The paper headers
%\markboth{Journal of \LaTeX\ Class Files,~Vol.~14, No.~8, August~2021}%
%{Shell \MakeLowercase{\textit{et al.}}: A Sample Article Using IEEEtran.cls for IEEE Journals}

%\IEEEpubid{0000--0000/00\$00.00~\copyright~2021 IEEE}
% Remember, if you use this you must call \IEEEpubidadjcol in the second
% column for its text to clear the IEEEpubid mark.

%\maketitle
\title{Comprehensive Review of \\ Deep Learning-Based 3D Point Cloud Completion Processing and Analysis}
%
%
% author names and IEEE memberships
% note positions of commas and nonbreaking spaces ( ~ ) LaTeX will not break
% a structure at a ~ so this keeps an author's name from being broken across
% two lines.
% use \thanks{} to gain access to the first footnote area
% a separate \thanks must be used for each paragraph as LaTeX2e's \thanks
% was not built to handle multiple paragraphs
%

\author{Ben Fei,
        Weidong Yang,
        Wen-Ming Chen,~\IEEEmembership{Member,~IEEE,}
        Zhijun Li,~\IEEEmembership{Senior Member,~IEEE,}
        Yikang Li,
        Tao Ma,
        Xing Hu and
        Lipeng Ma
        % <-this % stops a space
        
\thanks{This work was supported by the National Natural Science Foundation of China (U2033029)
\textit{Corresponding author: Wenming Chen and Weidong Yang}}

\thanks{Ben Fei, Weidong Yang and Lipeng Ma are with Shanghai Key Laboratory of Data Science, the School of Computer Science, Fudan University, Shanghai 200433, China (e-mail: bfei21@m.fudan.edu.cn; wdyang@fudan.edu.cn; lpma21@m.fudan.edu.cn).}% <-this % stops a space

\thanks{Wen-Ming Chen is with the Academy for Engineering and Technology, Fudan University, Shanghai 200433, China (e-mail: chenwm@fudan.edu.cn).}

\thanks{Zhijun Li is with the Department of Automation, University of Science and Technology of China, Hefei 230060, China (e-mail: zjli@ieee.org).}

\thanks{Yikang Li and Tao Ma is with Shanghai AI Laboratory, Shanghai, P. R. China (e-mail: liyikang@pjlab.org.cn; matao@pjlab.org.cn).}

\thanks{Xing Hu is with School of Optical-electrical Information and Computer Engineering, University of Shanghai for Science and Technology, 516 JunGong Road, Shanghai, P. R. China (e-mail: huxing@usst.edu.cn).}

\thanks{This article has supplementary downloadable material available at http://ieeexplore.ieee.org, provided by the authors}
}

\maketitle

\begin{abstract}
Point cloud completion is a generation and estimation issue derived from the partial point clouds, which plays a vital role in the applications of 3D computer vision. The progress of deep learning (DL) has impressively improved the capability and robustness of point cloud completion. However, the quality of completed point clouds is still needed to be further enhanced to meet the practical utilization. Therefore, this work aims to conduct a comprehensive survey on various methods, including point-based, view-based, convolution-based, graph-based, generative model-based, transformer-based approaches, etc. And this survey summarizes the comparisons among these methods to provoke further research insights. Besides, this review sums up the commonly used datasets and illustrates the applications of point cloud completion. Eventually, we also discussed possible research trends in this promptly expanding field.
\end{abstract}

%\def\abstractname{Note to Practitioners}
%\begin{abstract}
%The motivation of this paper is to summarize the application of point cloud completion in computer vision. In this paper, the existing point cloud completion algorithm is divided into different methods according to the network structure. And the characteristics and performance of these methods are systematically summarized. Moreover, the applications of point cloud completion in practical usages and its future development direction are summarized. In addition, this paper also summarizes the opportunities and challenges of deep learning in point cloud completion, in order to stimulate the potential development of deep learning. 
%\end{abstract}

% Note that keywords are not normally used for peerreview papers.

% For peer review papers, you can put extra information on the cover
% page as needed:
% \ifCLASSOPTIONpeerreview
% \begin{center} \bfseries EDICS Category: 3-BBND \end{center}
% \fi
%
% For peerreview papers, this IEEEtran command inserts a page break and
% creates the second title. It will be ignored for other modes.
\IEEEpeerreviewmaketitle

\section{Introduction}
% The very first letter is a 2 line initial drop letter followed
% by the rest of the first word in caps.
% 
% form to use if the first word consists of a single letter:
% \IEEEPARstart{A}{demo} file is ....
% 
% form to use if you need the single drop letter followed by
% normal text (unknown if ever used by the IEEE):
% \IEEEPARstart{A}{}demo file is ....
% 
% Some journals put the first two words in caps:
% \IEEEPARstart{T}{his demo} file is ....
% 
% Here we have the typical use of a "T" for an initial drop letter
% and "HIS" in caps to complete the first word.
\IEEEPARstart{W}{ith} the popularity of 3D scanning devices, including LiDAR, laser or RGB-D scanners, and so on, point clouds have become easier to capture and currently provoked a great deal of researches in the fields of robots, autonomous driving, 3D modeling and fabrication. However, the raw point clouds directly collected by these devices are primarily sparse and partial due to the occlusions, reflections, transparency, and restriction of devices' resolution and angles. Therefore, generating the complete point clouds from partial observations is crucial to boost the downstream applications. 

%in classification, identification, tracking, and shape manipulation applications.

The effectiveness of point cloud completion lies in its distinct and crucial role in various computer vision applications. \textbf{3D reconstruction}. The generation of complete 3D scenes is the foundation and important technology for numerous computer vision tasks, including 3D map reconstruction with high-resolution in autonomous driving, 3D reconstruction in robotics, and underground mining. For instance, point cloud completion in robot applications can help route planning and decision-making by constructing a 3D scene. Moreover, the large 3D environment reconstruction in underground mining space to accurately monitor mining safety. \textbf{3D detection}. The 3D object detection relies on complete point clouds to keep state-of-the-art (SOTA) performance. For example, cars in the distance captured by LiDAR tend to be sparse and often difficult to detect. It is noted that we often need to completely segment the target point cloud to complete it. For instance, when carrying the 3D detection on the KITTI dataset, the point clouds of vehicles need to be segmented and normalized to their scale and orientation before the complete point cloud of the vehicle can be obtained. \textbf{3D shape classification}. For 3D shape classification, the complete point clouds are ultimately needed by recovering from partial observations. The partial point cloud represents a small part of the object that is usually hard to recognize. Because point cloud completion plays an essential role in plenty of practical computer vision applications, there is an instant necessity for an extensive investigation of point cloud completion.

However, there are a few surveys on the completion of point cloud and downstream tasks, while the latest progress of deep learning in point cloud completion is urgently needed to be reviewed \cite{r1, r2, r3, r134, r135, r136, r137, r138, r139, r140}. To stimulate the developments of point cloud completion in industry and academy, we conducted a comprehensive review by summing up the rapid growth of point cloud completion technology in recent years (2017-2022), mainly including current deep-learning methods. Furthermore, we give comparisons among various deep-learning techniques.

\begin{figure}[ht]
    \centering
    \includegraphics[width=21pc]{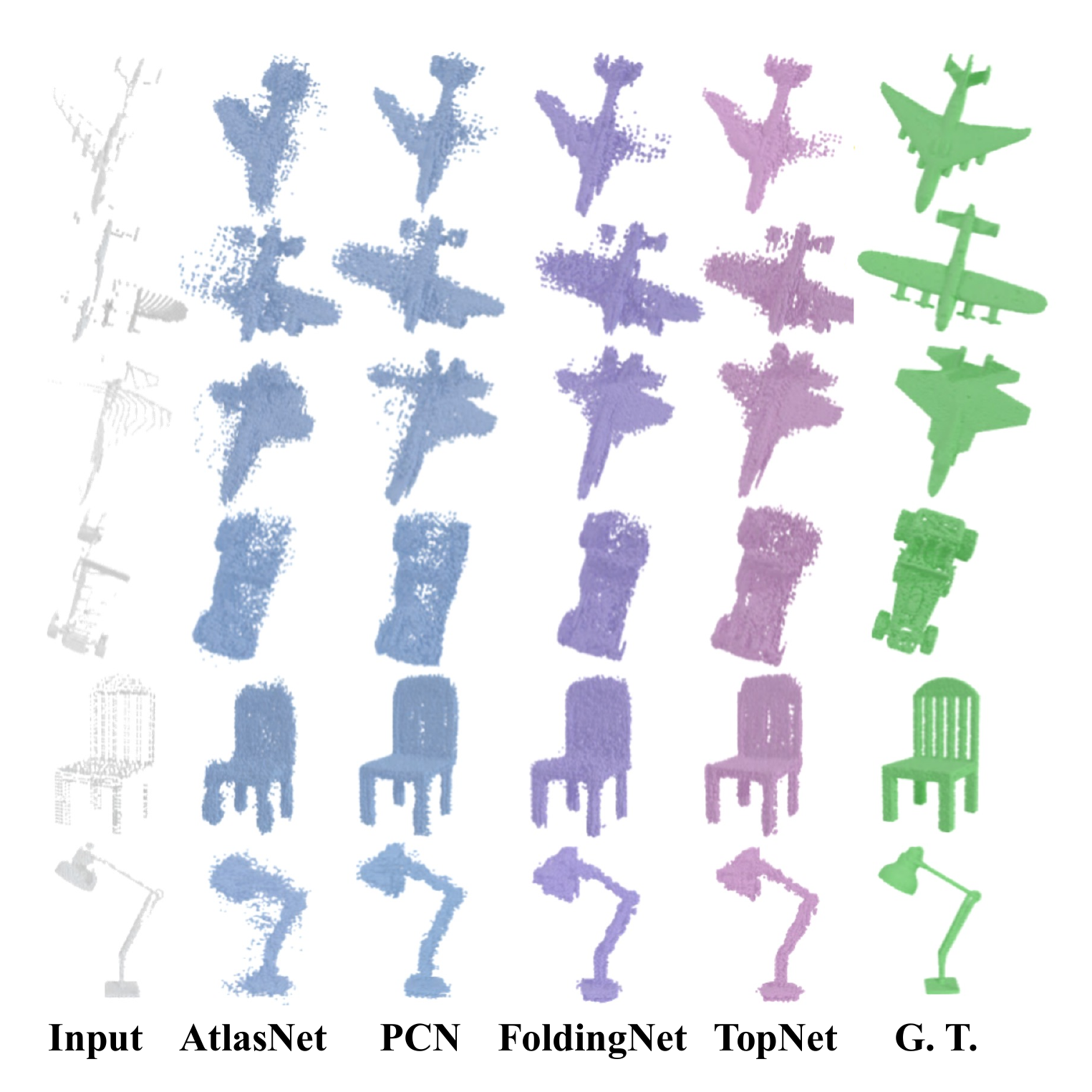}
    \caption{Schematic diagram of completion results of commonly used point cloud algorithms \cite{r59}.}
    %\label{fig:my_label}
    \vspace{-0.8cm}
\end{figure}

Over the past several years, researchers have tried numerous methods to address this issue in deep learning. Early attempts on point cloud completion \cite{r4, r5, r6, r7, r8, r9} tried to transfer mature methods from 2D completion tasks to 3D point clouds through voxelization and 3D convolution. However, these methods suffer from high computational costs with increasing spatial resolution. With the tremendous success of PointNet and PointNet++ \cite{r10, r11}, direct processing of 3D coordinates has become the mainstream of point cloud-based 3D analysis. This technique is further applied in many pioneering works of point cloud completion \cite{r12, r13, r14, r15, r16, r17, r57, r64}, where an encoder-decoder scheme is designed to produce complete point clouds. In recent years, many other methods, such as point-based, view-based, convolution-based, graph-based, generative model-based, and transformer-based methods, have been springing up and achieved significant results (Fig. 1).

\begin{figure*}[ht]
    \centering
    \includegraphics[width=43pc]{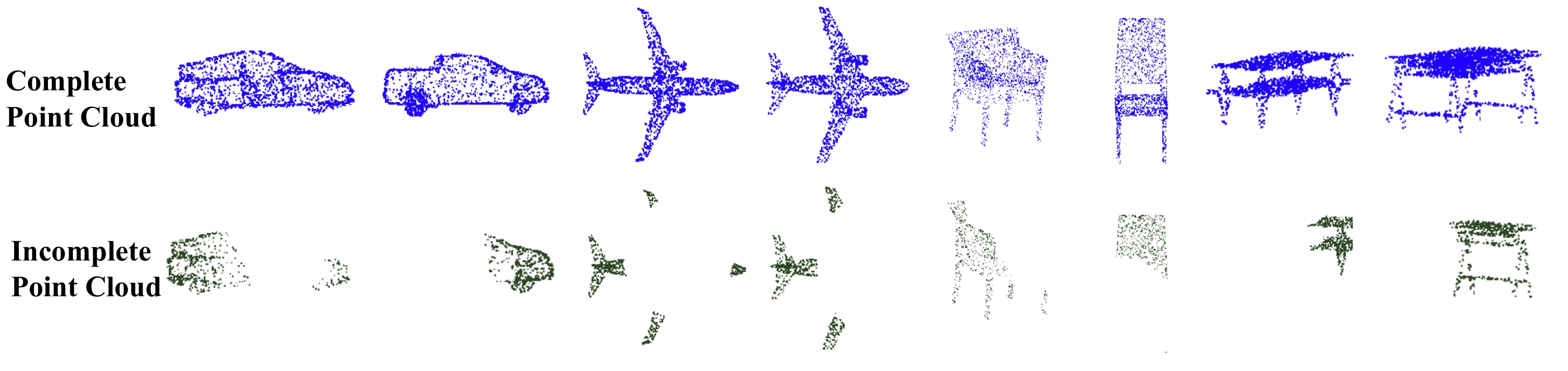}
    \caption{Schematic of complete point cloud and missing {70\%} point cloud.}
    %\label{fig:my_label}
    \vspace{-0.5cm}
\end{figure*}

Contrasted with the existing papers, the main contributions of the survey could be concluded as follows:
\begin{itemize}
	\item As far as we know, it is the first survey systematically covering nearly all DL methods on point cloud completion.
    \item This review introduces the latest and advanced progress in point cloud completion, together with their methods and contributions.
    \item Systematic comparisons of existing DL methods on a few public datasets, along with compact conclusions and profound discussions, are provided.
    \item We will discuss future research on DL-based point cloud completion at the end of this survey to stimulate improvement in this field.
\end{itemize}

\section{Reasons for incomplete point clouds}

\textbf{Definitions} A point cloud is a set of data points in space representing a 3D shape or an object. Point clouds are generally produced by 3D scanners or by photogrammetry software, consisting of a large number of points that geometrically represent the 3D surfaces of objects. A partial or incomplete point cloud means a point cloud that has missing points. The missing of points means a part of points in point clouds is missing for various reasons.

During data acquisition, the 3D laser scanner will be affected by the characteristics of the measured object, the processing method, and the environment, inevitably leading to the missing of points (Fig. 2). As shown in Fig. 3, the main reasons could be attributed to specular reflection, signal absorption, occlusion of external objects, self-occlusion of objects, and blind spots. The former two are due to the surface material of the objects, which might absorb or reflect the LiDAR signal in an unexpected way. The latter three are mainly due to occlusion, which can be completed with the aid of other parts of the objects or by utilizing multi-source data. Moreover, the stability of the 3D scanner in the scanning process also has a particular influence on the scanning quality.

After the data collection is completed, the point clouds also need to carry out a series of processing, such as point cloud denoising, smoothing, registration, and fusion. At the same time, these operations will significantly exacerbate the missing of points. This will not only affect the data integrity and lead to topology errors but also affect the quality of the point cloud refactoring, the 3D model reconstruction, local spatial information extraction, and subsequent processing.

\begin{figure}[ht]
    \centering
    \includegraphics[width=20pc]{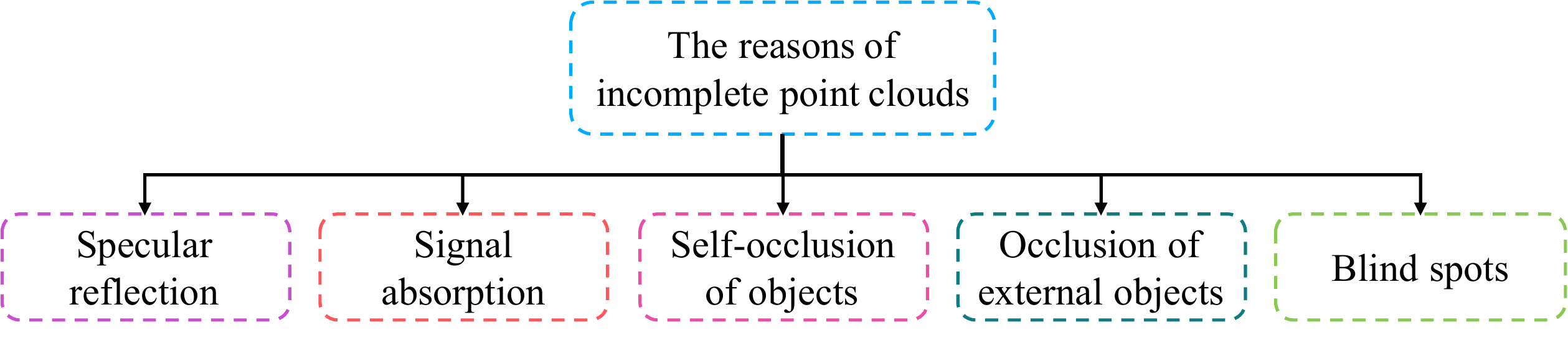}
    \caption{Reasons for incomplete point clouds.}
    %\label{fig:my_label}
    \vspace{-0.5cm}
\end{figure}

\section{Challenges}

\subsection{Structural information challenges}

The reconstruction of a complete point cloud is challenging because the structural information required for the task of point cloud completion runs counter to the disordered and unstructured nature of the point clouds. 3D object point clouds in the real world could be regarded as low-level and high-level configurations, including surfaces, semantic parts, geometrical elements, etc. Existing point cloud generation frameworks either exclude structure in their devised solutions or assume and execute a particular structure/topology for generating the complete point clouds of 3D objects, for example, a set of surfaces or manifolds. Hence, learning the structural features of the point cloud becomes critical for a better complete point cloud.

% Up to now, the ViPC \cite{r41} has performed best on using the points of another part (such as fine parts) as completion constraints to infer the local point distribution of a part (such as coarse parts) and make full use of structural information. This enables the network to retain the local details from the input incomplete point cloud, resulting in a better and plausible completion.

\subsection{Fine-grained complete shapes challenges}

3D shape completion is supposed to reconstruct a reasonable fine-grained complete point cloud using relational structure information, such as geometrical symmetry, regular arrangements, and surface smoothness. Although several works have been fully exploiting the structure information through iterative refinement \cite{r72}, integration of global features and local features \cite{r48}, skip-connections \cite{r58}, residual connections \cite{r22}, etc., more efforts should be paid to generating the fine-grained complete shapes.

% For instance, CRN \cite{r72} designed an iterative refinement decoder along with a contraction of features and extension unit for the refinements of positions. MSN \cite{r22} merged the coarse output with the input, and learned a point-wise residual for the combination. ECG \cite{r48} integrated global features with local ones by the generation of skeleton and refinement of details .

Therefore, this review will investigate the SOTA completion's performance and discuss the solutions they used in tackling these two significant challenges.

% Please add the following required packages to your document preamble:
% \usepackage{graphicx}
\begin{table*}[htbp]\footnotesize
\centering
\setlength\tabcolsep{1pt}
\caption{Summary of existing datasets for point cloud completion.}
{%
\begin{tabular}{c|ccccccc}
\hline
Name                                                                            & Year & Classes  & Sensors or origin & Type           & Views  & Resolutions            & Description                                                                                                                                                                    \\ \hline
PCN  \cite{r18}                                                                             & 2015 & 8        & CAD               & Synthetic      & 8      & \begin{tabular}[c]{@{}c@{}}2048/4096/\\  8192/16384 \end{tabular}           & Derived from ShapeNet.                                                                                                                                                         \\ \hline
ShapeNet55  \cite{r89}                                                                    & 2021 & 55       & CAD               & Synthetic      & all possible views & 8192 & \begin{tabular}[c]{@{}c@{}}Contains all the objects in ShapeNet\\  from 55 categories.\end{tabular}                                                                            \\ \hline
ShapeNet34  \cite{r89}                                                                    & 2021 & 34       & CAD               & Synthetic      & all possible views & 8192 & \begin{tabular}[c]{@{}c@{}}Contains 21 unseen categories and\\  34 seen categories.\end{tabular}                                                                               \\ \hline
KITTI     \cite{r19}                                                                      & 2012 & 8        & RGB \& LiDAR      & Urban (Driving) & -     & \begin{tabular}[c]{@{}c@{}}depends on\\  the network\end{tabular}              & \begin{tabular}[c]{@{}c@{}}Derived from KITTI, real-world  \\ LiDAR scans. Sparse in nature.\end{tabular}                                                                       \\ \hline
ModelNet     \cite{r20}                                                                   & 2015 & 10 or 40 & CAD               & Synthetic      & 12         &  2048/16384      & Proposed by princeton.                                                                                                                                                         \\ \hline
Completion3D   \cite{r16}                                                                 & 2019 & 8        & CAD               & Synthetic      & -      &   1024/2048/16384         & \begin{tabular}[c]{@{}c@{}}Derived from ShapeNet. The partial  \\3D shapes are generated by  \\  back-projected 2.5D depth images from \\ partial views into 3D space.\end{tabular} \\ \hline
\begin{tabular}[c]{@{}c@{}}Multi-View Partial \\ Point Cloud (MVP) \cite{r91} \end{tabular} & 2021 & 16       & CAD               & Synthetic      & 26   & 2048/4096/16384              & \begin{tabular}[c]{@{}c@{}}Diversity of uniform views; Large-scale \\ and high-quality; Rich categories.\end{tabular}                                                          \\ \hline
\begin{tabular}[c]{@{}c@{}}Single-View Point \\ Cloud Completion \cite{r91} \end{tabular}   & 2021 & 16       & CAD               & Synthetic      & 1         & 2048/4096/16384        & A large subset of MVP                                                                            \\ \hline
\end{tabular}%
}
\vspace{-0.5cm}
\end{table*}

\begin{figure*}[htbp]
    \centering
    \includegraphics[width=35pc]{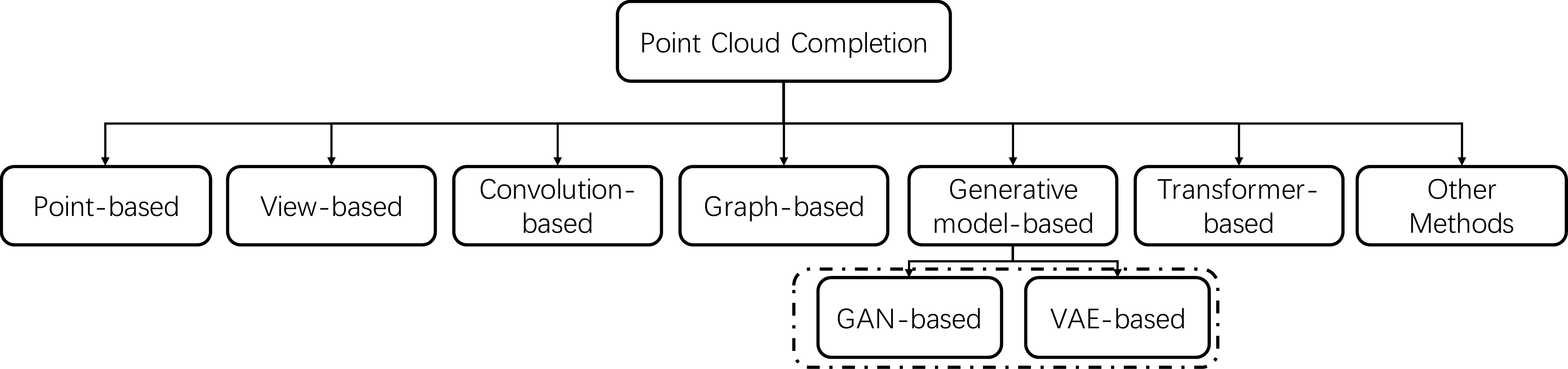}
    \caption{Taxonomy of point cloud completion}
    %\label{fig:my_label}
    \vspace{-0.5cm}
\end{figure*}

\section{Datasets}

\begin{figure*}[htp]
    \centering
    \includegraphics[width=43pc]{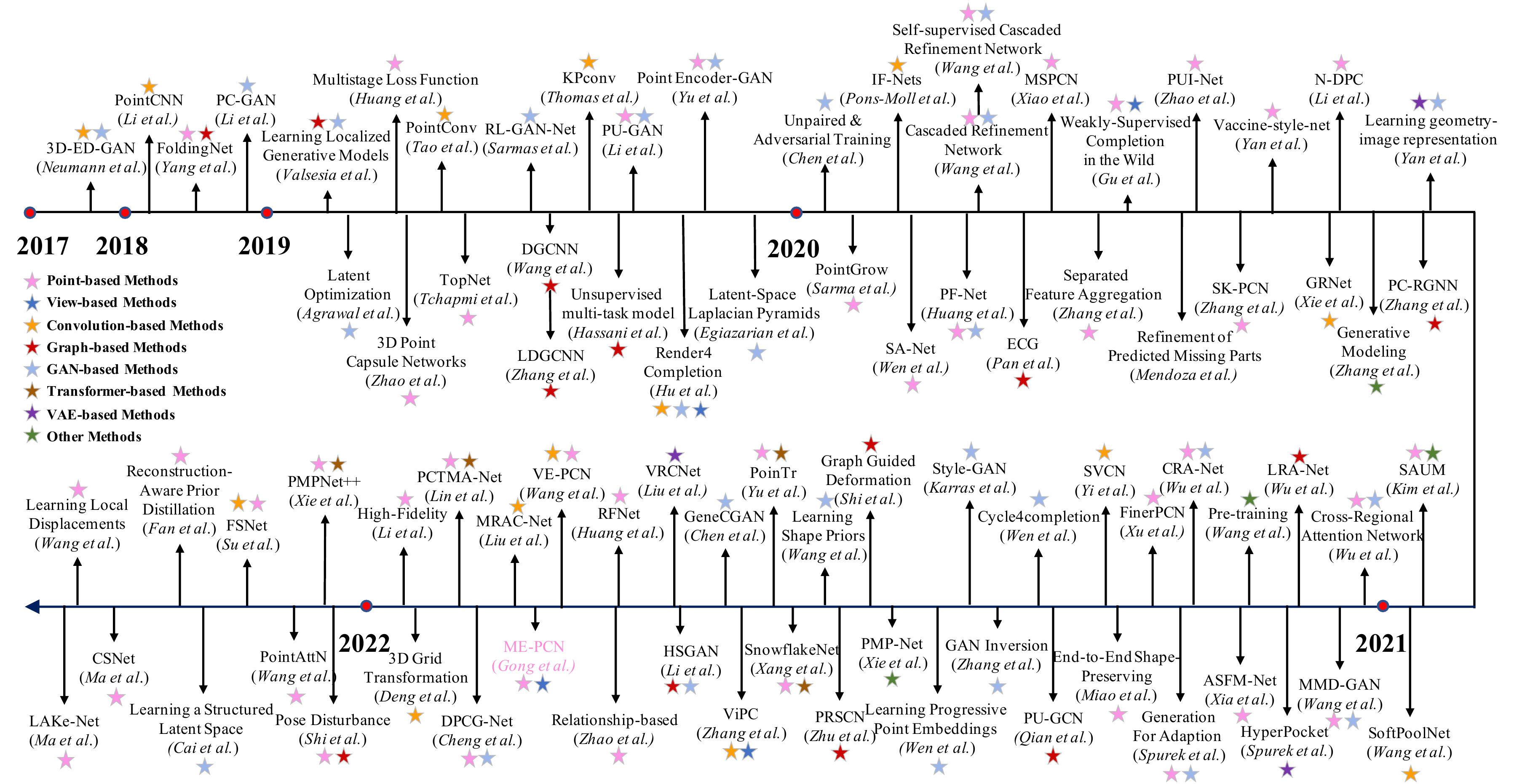}
    \caption{Chronological summarisztion of the recent and most relevant DL-based methods for point cloud completion.}
    \vspace{-0.5cm}
    %\label{fig:my_label}
\end{figure*}

As for 3D shape completion, the datasets could be categorized into two types: artificial datasets and real-world datasets (Table I). The most studied four datasets are as follows:

% ShapeNet \cite{r18}: The Computer-Aided Design (CAD) dataset derived from PCN \cite{r17}, totally containing 30974 3D models from 8 specific categories. The ground truth point clouds consist of 16384 uniformly sampled on the surfaces.
PCN \cite{r18}: The Computer-Aided Design (CAD) dataset derived from PCN \cite{r17}, totally containing 30974 3D models from 8 specific categories. The ground truth point clouds consist of 16384 uniformly sampled on the surfaces.

% KITTI \cite{r19}: The dataset is collected through a Velodyne laser scanner. The odometry dataset was originally designed to evaluate the performance of stereo matching, which consists of LiDAR point clouds, stereo sequences, and ground truth poses. It contains 22 stereo sequences, among which the training set consists of 11 sequences (id 00-10) with ground-truth trajectories, while the evaluation set contains 11 sequences (id 11-21) without ground truth.
KITTI \cite{r19}: The dataset is collected through a Velodyne laser scanner. The odometry dataset was originally designed to evaluate the performance of stereo matching, which contains LiDAR point clouds of 22 stereo sequences. The partial cars were collected to evaluate the performance of point cloud completion methods in real-world scans in the absence of ground truth.

ModelNet40 \cite{r20}: A comprehensive set of 3D CAD models. Its objects consist of 40 categories and 13356 models.

Completion3D \cite{r16}: An online platform for evaluating shape-completion methods on the basis of a subset derived from the ShapeNet dataset. It is noted that some paired point clouds face the problem of unmatched scales, which can be tackled by the one-side CD loss proposed in \cite{r141}.

In addition to the above datasets, the Shapenet 34/55 \cite{r89} and MVP datasets \cite{r91} have recently been proposed to increase the variety and number of objects, diverse viewpoints, and varying degrees of defects as close as possible to real-world objects.

\section{Metrics}
For 3D point cloud completion, the Chamfer Distance (CD) \cite{r117} and Earth Mover’s Distance (EMD) \cite{r117} are the most frequently used performance criteria. CD tries to find the minimum distance between two sets of points, while EMD evaluates the reconstruction quality of the point clouds.

(1) Chamfer distance (CD)

\begin{equation}\footnotesize
\setlength\abovedisplayskip{1pt}
\setlength\belowdisplayskip{1pt}
CD(\textbf{S}_1,\textbf{S}_2)={\dfrac{1}{|\textbf{S}_1|}}\sum_{x\in{\textbf{S}_1}}{\underset{y\in{\textbf{S}_2}}{min}{\parallel{x-y}\parallel_2}} +
\\{\dfrac{1}{|\textbf{S}_2|}}\sum_{y\in{\textbf{S}_2}}{\underset{x\in{\textbf{S}_1}}{min}{\parallel{y-x}\parallel_2}}
\end{equation}

By definition, CD represents the sum of the average closest distance from points in the output $S_1$ to points in the complete point clouds $S_2$, and from points in $S_2$ to points in $S_1$. There are two variants to CD: CD-T (CD-${\ell}_1$) and CD-P (CD-${\ell}_2$). The definitions of CD-T and CD-P between two point clouds $S_1$ and $S_2$ are as follows:

\begin{equation}
\setlength\abovedisplayskip{1pt}
\setlength\belowdisplayskip{1pt}
\begin{aligned}
\setlength\abovedisplayskip{1pt}
\setlength\belowdisplayskip{1pt}
&\mathcal{L}_{\textbf{S}_1, \textbf{S}_2}=\frac{1}{|\mathbf{S}_{1}|} \sum_{x \in \textbf{S}_1} \min _{y \in \textbf{S}_2}\|y-x\|_{2}^{2} \\
&\mathcal{L}_{\textbf{S}_2, \textbf{S}_1}=\frac{1}{|\mathbf{S}_{2}|} \sum_{y \in \textbf{S}_2} \min _{x \in \textbf{S}_1}\|x-y\|_{2}^{2} \\
&\mathcal{L}_{C D-T}(\textbf{S}_1, \textbf{S}_2)=\mathcal{L}_{\textbf{S}_1, \textbf{S}_2}+\mathcal{L}_{\textbf{S}_2, \textbf{S}_1} \\
&\mathcal{L}_{C D-P}(\textbf{S}_1, \textbf{S}_2)=\left(\sqrt{\mathcal{L}_{\textbf{S}_1, \textbf{S}_2}}+\sqrt{\mathcal{L}_{\textbf{S}_2, \textbf{S}_1}}\right) / 2
\end{aligned}
\end{equation}

(2) Earth Mover’s Distance (EMD)

EMD aims to find out a bijection $\phi : S_1  \to S_2$ to minimize the average distance between corresponding points from partial and complete ones. Different from CD, the size of $S_1$ and $S_2$ needs to be same.
\begin{equation}
\setlength\abovedisplayskip{1pt}
\setlength\belowdisplayskip{1pt}
EMD(\textbf{S}_1,\textbf{S}_2)={\underset{{\phi:{\textbf{S}_1}\to{\textbf{S}_2}}}{min}}{\frac{1}{|\mathbf{S}_{1}|}}\sum_{x\in{\textbf{S}_1}}{{\parallel{x-\phi{(x)}}\parallel_2}}
\end{equation}

(3) Fidelity error (FD), Maximum Mean Discrepancy (MMD) and consistency

Fidelity error (FD), Consistency, and minimal matching distance (MMD) are proposed by PCN as evaluation metrics \cite{r17}. Fidelity is utilized to measure how well the input is preserved, which calculates the average distance between the point in the input and the corresponding nearest neighbor in the output. MMD is used to measure how much the model's outputs reconstruct a typical car. Consistency aims to estimate how consistent the model's outputs are against variations in the inputs.

(4) Density-aware Chamfer Distance (DCD)

DCD \cite{r125} is derived from CD and it can detect disparity of density distributions. DCD pays attention to both the overall structure and local geometric details.

%\begin{equation}
%\begin{split}
%DCD(\textbf{S}_1,\textbf{S}_2)={\dfrac{1}{2}}({\dfrac{1}{|\textbf{S}_1|}}{\underset{x\in{\textbf{S}_1}}{\sum}}{(1-{\dfrac{1}{n_\hat{y}}}{e^{-\alpha\parallel{x-\hat{y}}\parallel_2}}} + \\
%({\dfrac{1}{|\textbf{S}_2|}}{\underset{y\in{\textbf{S}_2}}{\sum}}{(1-{\dfrac{1}{n_\hat{x}}}{e^{-\alpha\parallel{\hat{x}-y}\parallel_2}}}))
%\end{split}
%\end{equation}

\begin{equation}
\setlength\abovedisplayskip{1pt}
\setlength\belowdisplayskip{1pt}
\begin{array}{r}
\operatorname{DCD}\left(\mathbf{S}_{1}, \mathbf{S}_{2}\right)=\frac{1}{2}\left(\frac { 1 } { | \mathbf { S } _ { 1 } | } \sum _ { x \in \mathbf { S } _ { 1 } } \left(1-\frac{1}{n_{\hat{y}}} e^{-\alpha\|x-\hat{y}\|_{2}}+\right.\right. \\
\left(\frac{1}{\left|\mathbf{S}_{2}\right|} \sum_{y \in \mathbf{S}_{2}}\left(1-\frac{1}{n_{\hat{x}}} e^{-\alpha\|\hat{x}-y\|_{2}}\right)\right)
\end{array}
\end{equation}

(5) F-Score

The F-score proposed by Tatarchenko et al. \cite{r129} is to evaluate the distance between the surfaces of the object and is treated as the harmonic mean between precision and recall. The precision counts the percentage of reconstructed points within a certain distance to the ground truth, which stands for the accuracy of the reconstruction. On the other hand, the recall counts the percentage of points on the ground truth within a certain distance to the reconstruction, representing the completeness of the reconstruction. The distance threshold $d$ can be leveraged to control the strictness of the F-score. The F-score can evaluate the percentage of points or surface area that are reconstructed correctly, which can be defined as follows:

\begin{equation}
    \setlength\abovedisplayskip{1pt}
    \setlength\belowdisplayskip{1pt}
    \mathrm{F}-\operatorname{Score}(d)=\frac{2 P(d) R(d)}{P(d)+R(d)}
\end{equation}

where $P(d)$ and $R(d)$ denote the precision and recall for a distance threshold $d$, respectively.

\begin{equation}
    \setlength\abovedisplayskip{1pt}
    \setlength\belowdisplayskip{1pt}
    P(d)=\frac{1}{|\mathbf{S}_{1}|} \sum_{r \in \mathcal{S}_1}\left[\min _{t \in \mathcal{S}_2}\|t-r\|<d\right]
\end{equation}
\begin{equation}
    \setlength\abovedisplayskip{1pt}
    \setlength\belowdisplayskip{1pt}
    R(d)=\frac{1}{|\mathbf{S}_{2}|} \sum_{t \in \mathcal{S}_2}\left[\min _{r \in \mathcal{S}_1}\|t-r\|<d\right]
\end{equation}

where $\mathcal{S}_1$ is a reconstructed point set being evaluated and $\mathcal{S}_2$ is the ground truth. $|\mathbf{S}_{1}|$ and $|\mathbf{S}_{2}|$ are the numbers of points of $\mathcal{S}_1$ and $\mathcal{S}_2$, respectively.

(6) Uniformity

Uniformity \cite{r77} is usually employed to evaluate the distribution uniformity of the completed point clouds, which can be formulated as:

\begin{equation}
    \setlength\abovedisplayskip{1pt}
    \setlength\belowdisplayskip{1pt}
    \text { Uniformity }(p)=\frac{1}{M} \sum_{i=1}^{M} \mathrm{U}_{\text {imbalance}}\left(Q_{i}\right) \mathrm{U}_{\text {clutter}}\left(Q_{i}\right)
\end{equation}

where $Q_{i}(i=1,2, \ldots, M)$ is a point subset cropped from a patch of the output $\mathcal{S}_1$ using the farthest sampling and ball query of radius $\sqrt{p}$. The term $\mathrm{U}_{\text {imbalance }}$ and $U_{\text {clutter}}$ account for the global and local distribution uniformity, respectively.

\begin{equation}
    \setlength\abovedisplayskip{1pt}
    \setlength\belowdisplayskip{1pt}
    \mathrm{U}_{\text {imbalance }}\left(Q_{i}\right)=\frac{\left(\left|Q_{i}\right|-\hat{n}\right)^{2}}{\hat{n}}
\end{equation}

where $\hat{n}=p|\mathcal{S}_1|$ is the expected number of points in $Q_{i}$.

\begin{equation}
    \setlength\abovedisplayskip{1pt}
    \setlength\belowdisplayskip{1pt}
    \mathrm{U}_{\text {clutter }}\left(Q_{i}\right)=\frac{1}{\left|Q_{i}\right|} \sum_{j=1}^{\left|Q_{i}\right|} \frac{\left(d_{i, j}-\hat{d}\right)^{2}}{\hat{d}}
\end{equation}

where $d_{i, j}$ represents the distance to the nearest neighbor for the $j$-th point in $Q_{i}$, and $\hat{d}$ is roughly $\sqrt{\frac{2 \pi p}{\left|Q_{i}\right| \sqrt{3}}}$ if $Q_{i}$ has a uniform distribution.

\section{Methods}

As shown in Fig. 4, based on the network structure employed in point cloud completion and generation, the existing architectures could be categorized into point-based, view-based, convolution-based, graph-based, transformer-based, generative model-based, and other methods. Nearly all milestone contributions are clearly illustrated in Fig. 5. Since most works are hybrid methods, they might belong to several methods according to their stated highlights. Therefore, we will discuss these works if they utilized the methods that can be categorized into these classes.

\subsection{Point-based methods}

The point-based approaches usually model every point independently by utilizing Multi-layer Perceptrons (MLPs). The global feature is then aggregated through a symmetric function (such as Max-Pooling) because of the transformation invariance of the point clouds. Whereas the geometric information and correlations in the whole point group are still not entirely considered. As a commonly used method for processing the features, we only review the methods mainly using point-based networks in this section.

\textbf{Preliminary works} Pioneered by PointNet \cite{r10}, a few works used MLP for the processing and recovering of the point clouds due to its concise and non-negligible ability of representations \cite{r89, r71 ,r68,r77}. PointNet++ \cite{r11} and TopNet \cite{r16} incorporated a hierarchical structure to take the geometric information into consideration. PointNet++ proposed two set abstraction layers, which intelligently aggregate multi-level information, while TopNet proposed a new decoder that generates a structured point cloud without supposing any particular structure or topology. Inspired by PointNet and PointNet++, Yu et al. \cite{r21} proposed PU-Net learn multi-scale features by feature scaling based on sub-pixel convolutional layers. The scaling restoration method performs convolution with 1x1 kernels on extracted features. Then, the extracted features are decomposed and reconstructed into a cluster of up-sampling points. And the joint loss function is utilized to distribute the generated point cloud on the potential surface evenly. Nevertheless, PU-Net is primarily designed to generate a single denser point cloud from a sparse point cloud rather than perform point cloud completion. It cannot fill large holes and missing parts, nor can it add meaningful points to heavily down-sampled parts of the point cloud.

To mitigate the structure loss brought by MLP, the proposed AtlasNet \cite{r12}, and MSN \cite{r22} reconstruct the complete output through evaluating a set of parametric surface elements, from which the complete point clouds could be generated. Specifically, AtlasNet \cite{r12} took an additional input of a 2D point in the unit square and applied it to produce a single point on the surface. Thus, the output is a continuous image of a plane. This method could be reiterated many times to reconstruct a 3D shape from a combination of numerous surface elements. For avoiding structural loss, MSN \cite{r22} introduced the morphing-based decoder that can morph the unit squares as a set of surface elements aggregated into the coarse point cloud.

\textbf{PCN-drived methods} For the first time, Hebert et al.\cite{r17} proposed a learning-based shape completion method, Point Completion Network (PCN). Unlike existing approaches, PCN straightly works on original point clouds and does not require any assumption of structural (such as symmetry) or annotation about the underlying shape (such as semantic class). It features a decoder design that allows the generation of fine-grained completions while maintaining a small number of parameters. By combining with PCN and point-wise convolution, Xu et al. \cite{r23} devised FinerPCN to generate complete and fine point clouds in a coarse-to-fine manner by taking into account local information. After that, Zhang et al. \cite{r24} proposed a skeleton-bridged point completion network (SK-PCN). The SK-PCN features a 3D skeleton that is predicted to learn global information. Following that, the surface is completed by using displacements with skeletal points. In MSPCN, Xiao et al. \cite{r25} used a tandem of up-sampling modules to reconstruct fine-grained output and supervised each stage to generate outputs with more information and beneficial intermediate for the following phase. Furthermore, they proposed a method to identify critical sets (MVCS) for supervision by combining selected points with max-pooling and volume-down-sampling points. This MVCS could take the vital features and the overall shape into consideration. By regarding partial point clouds as different samples of the same distribution, the Structure Retrieval based Point Completion Network (SRPCN) \cite{r140} was devised. SRPCN uses k-means clustering to extract structure points and disperses them into distributions, and then KL Divergence is used as a metric to find the complete structure point cloud that best matches the input in a database. To obtain a good feature representation that can capture both global structure and local geometric details, FSNet \cite{r136} was presented to adaptively aggregate point-wise features into a 2D structured feature map by learning multiple latent patterns from local regions. Then, FSNet is integrated into a coarse-to-fine pipeline for point cloud completion.

\begin{figure}[ht]
    \centering
    \includegraphics[width=21pc]{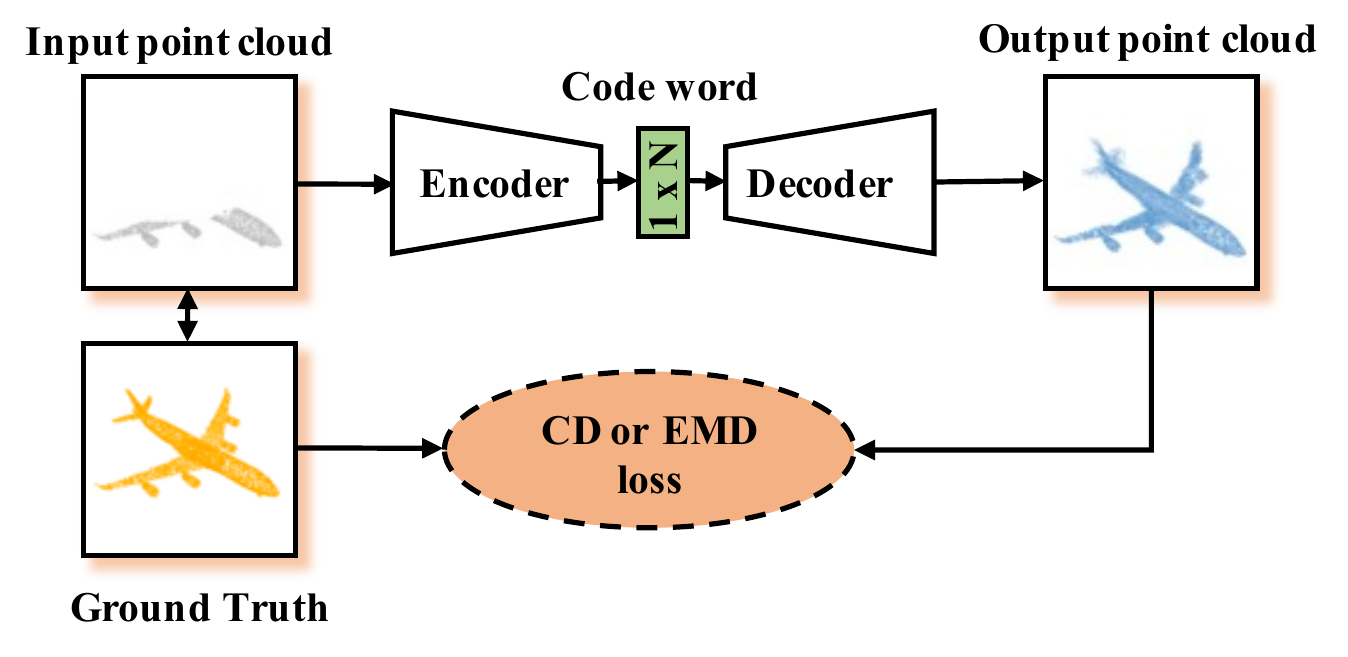}
    \caption{The illustration of end-to-end network for point clouds completion. $N$ represents the dimension of latent space.}
    \vspace{-0.5cm}
\end{figure}

\begin{figure*}[ht]
    \centering
    \includegraphics[width=43pc]{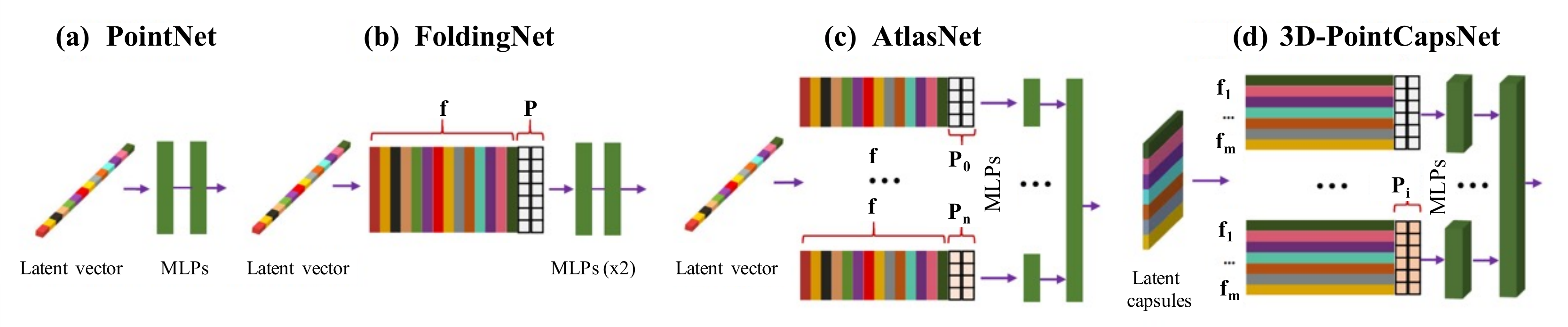}
    \caption{Comparison of four different SOTA 3D point decoders. (a) PointNet\cite{r10} utilizes a single latent vector without surface assumption. (b) FoldingNet \cite{r55} learns a 1D latent vector together with a fixed 2D grid. (c) The AtlasNet \cite{r12} learns a deformation to transform multiple 2D configurations into local 2-manifolds. (d) The point-capsule network \cite{r28} can learn multiple latent representations, each of which can fold a distinct 2D grid onto a specific local patch.}
    %\label{fig:my_label}
    \vspace{-0.5cm}
\end{figure*}

\begin{figure}[htbp]
    \centering
    \includegraphics[width=21pc]{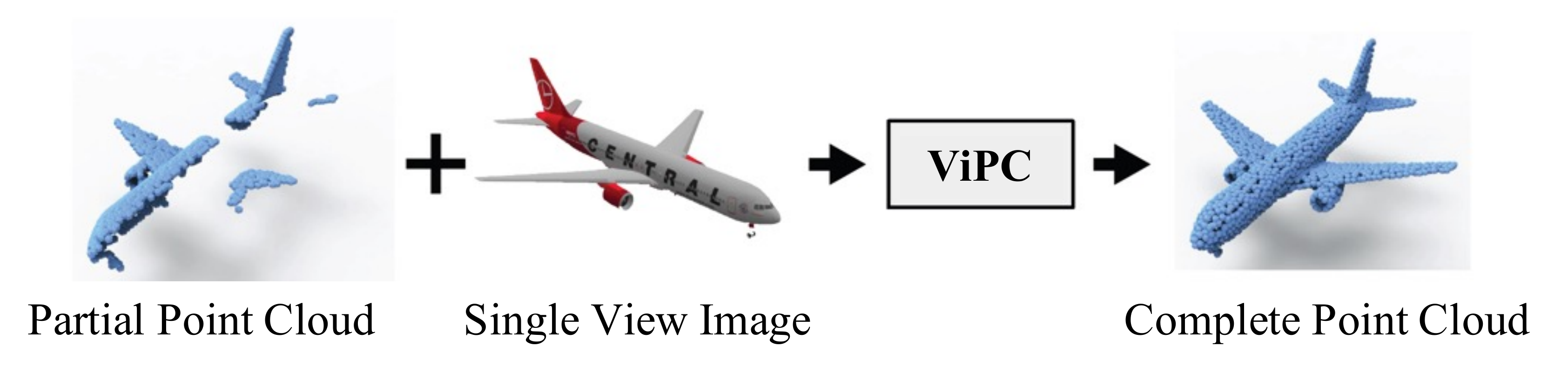}
    \caption{ViPC is a method to complete a partial point cloud using complementary information from additional single-view images \cite{r44}.}
    \vspace{-0.5cm}
\end{figure}

\textbf{End-to-end mechanism} In point-based methods, the end-to-end manner is widely used in the network architecture. In the encoder-decoder scheme (Fig. 6), the encoder in completion architecture aims at extracting both the global 3D shape features and the regional features of each point. At the same time, the decoder generates a completion point cloud and refines it. Stilla et al. \cite{r30} devised an S2UNet network to reconstruct a more uniform and fine-grained structure from a sparse point cloud in the applications of vehicles in an end-to-end manner. Significantly, they adopt an up-sampling approach to generate a more uniform point cloud. Furthermore, they devised ASFM-Net \cite{r31}, in which the asymmetrical Siamese auto-encoder (AE) generates a coarse but complete output, and the following refinement unit aims to recover a final point cloud with fine-grained details. Mendoza et al. \cite{r32} proposed a network with an end-to-end pattern consisting of two neural networks: the missing part prediction network and the merging-refinement network. This method predicts and integrates missing parts while preserving the existing geometry and refining the details. Miao et al. \cite{r33} presented a shape-preserving completion network to maintain the 3D shape and recover the fine-grained information of the reconstructed 3D shapes by designing an encoder-decoder scheme. This shape-preserving network could learn the global features and integrate regional information of adjacent points with various orientations and scales. During the decoding process, the information would be fused into latent vectors. Liao et al. \cite{r34} proposed a sparse-to-dense multi-encoder neural network (SDME-NET) in an end-to-end manner for completion while preserving details of the 3D shape. Notably, the defect point cloud would be completed and refined in two stages, from sparse to dense. In the first phase, they generated coarse but complete results based on a two-layer PointNet. In the second phase, they leveraged the sparse results of the first stage using PointNet++ to yield a high-density and high-fidelity point cloud. LAKe-Net \cite{r139} was proposed as a topology-aware point cloud completion model by localizing aligned keypoints, with a novel Keypoints-Skeleton-Shape prediction manner. Cai et al. \cite{r134} aim to learn a unified and structured latent space that encodes both partial and complete point clouds to improve partial-complete geometry consistency in an unsupervised manner. Moreover, they applied tailored structured latent supervisions between a series of related partial point clouds to enhance the learning of the structured latent space. In this way, they could reconstruct more accurate complete point clouds equipped with better fine-grained shape details. To reduce the dependency on paired data when training a point cloud completion model, RaPD \cite{r135} was proposed as a novel semi-supervised method. A two-stage training scheme is introduced in RaPD. The first stage learns robust semantic prior through reconstruction-aware pre-training, while the second stage learns the final completion deep model by prior distillation and self-supervised completion learning. CS-Net \cite{r137} is an end-to-end network to complete the point clouds contaminated by noises or containing outliers. In CS-Net, the completion and segmentation modules work collaboratively to promote each other and benefit from the designed cascaded structure. With the help of the segmentation, the more clean point cloud is fed into the completion module. Wang et al. \cite{r138} proposed an architecture that relies on three layers used successively within an encoder-decoder structure, attaining significant improvements producing high resolutions reconstruction with fine-grained details. The first one carries out feature extraction by matching the point features to a set of pre-trained local descriptors. Then, to avoid losing individual descriptors as part of standard operations such as max-pooling, an alternative neighbor-pooling operation is proposed that relies on adopting the feature vectors with the highest activations. Finally, upsampling in the decoder modifies our feature extraction in order to increase the output dimension.

Moreover, two feature assemble strategies were proposed to exploit the function of multi-scale features and integrate different information to represent given parts and missing parts, respectively. The global and local feature aggregation (GLFA) and the residual feature aggregation (RFA) were dubbed \cite{r35}. These two approaches represent the two types of features and recover coordinates with the help of their combination \cite{r35}. Furthermore, a refinement module was also designed to prevent uneven distribution of the generated point cloud.

Given the scenes composed of many objects, Zhao et al. \cite{r115} devised a partial point cloud completion approach, which mainly emphasizes paired scenarios where two objects are very close and contextually related. And a network was designed to encode the geometry of the individual shapes and the spatial relations between different objects from paired scenes. The two-path scheme is monitored by the consistency loss between different completion sequences by the merits of conditional completion. This approach could deal with the complicated case of objects severely occluding each other.

To tackle the challenging dense 3D point cloud completion problem, Li et al. \cite{r124} proposed a framework to perform end-to-end low-resolution recovery first, followed by a patch-wise noise aware upsampling. This method decodes a complete but sparse shape, followed by iterative refinement. Then the point cloud goes through preserving trustworthy information by symmetrization and patch-wise up-sampling. In this way, a high-fidelity dense point cloud can be achieved. Recently, a Recurrent Forward Network (RFNet) consisting of three modules (Recurrent Feature Extraction (RFE) module, Forward Dense Completion (FDC) module, and Raw Shape Protection (RSP) module) was devised \cite{r114}. The RFE extracts multiple global features from the incomplete point clouds for different recurrent levels, while the FDC produces the output in a coarse-to-fine pipeline. Further, the RSP introduces details from the original incomplete shapes to refine the completion results. In addition, a Sampling Chamfer Distance was proposed to capture the shapes of objects, and a Balanced Expansion Constraint was devised to limit the expansion distances from coarse to fine.

%\begin{itemize}
%    \item However, the information loss of structure details on local regions is the critical challenge in the encoding process, which should be preserved.
%\end{itemize}

\textbf{Attention-assited methods} Attention is a flexible mechanism for learning information self-adaptively, and the accumulated important information is weighted highly. By maintaining the spatial arrangements of the partial point clouds, the 3D point-capsule network \cite{r28} was devised to handle the sparse 3D point clouds with an auto-encoder. The creation of a 3D capsule network results from the unified, universal 3D auto-encoders. As shown in Fig. 7, the capsule network chooses a promising direction, where a large number of convolution filters realize the learning of the capsule set through dynamic routing. Be integrated with an encoder-decoder architecture, and the PUI-Net \cite{r29} has the advantage of extracting features with several cascaded Attention Conv Units and concatenating multi-level features before expansion. By using the extracted discriminative features, the dense feature map of the fine-grained point cloud is generated by a non-regional feature expansion unit. Li et al. \cite{r26} presented a dense point cloud completion model (N-DPC), combining self-attention units with the fusion of local features and global features. Sun et al. \cite{r27} proposed an auto-regressive network PointGrow with self-attention, which operates recurrently. PointGrow sampled each point according to a conditional distribution given its previously-generated points, allowing inter-point correlations to be well-exploited. PointAttN \cite{r130} employs the leverage the cross-attention and self-attention mechanisms to tackle the point completion task in a coarse-to-fine manner. It mainly comprises three modules: a feature extractor block for local geometric structure and global shape feature capturing, a seed generator block for the coarse point cloud generation, and a point generator block to produce the fine-grained point cloud.

\textbf{Folding-derived methods}
As a generic architecture firstly proved by Yang et al. \cite{r55}, a Folding-based decoder can reconstruct arbitrary point clouds from 2D grids for objects with detailed structures with low reconstruction errors (Fig. 9, Fig. 10). The FoldingNet is like applying a "virtual force" that deforms/cuts/stretches a 2D grid lattice onto a 3D surface. This deforming force should be affected or modulated by the interconnections induced by the adjacent meshes. Due to the intermediate folding steps and training processes in the decoder could be represented by reconstruction points, the gradual variation of folding force could be seen intuitively.

\begin{figure}[ht]
    \centering
    \includegraphics[width=12pc]{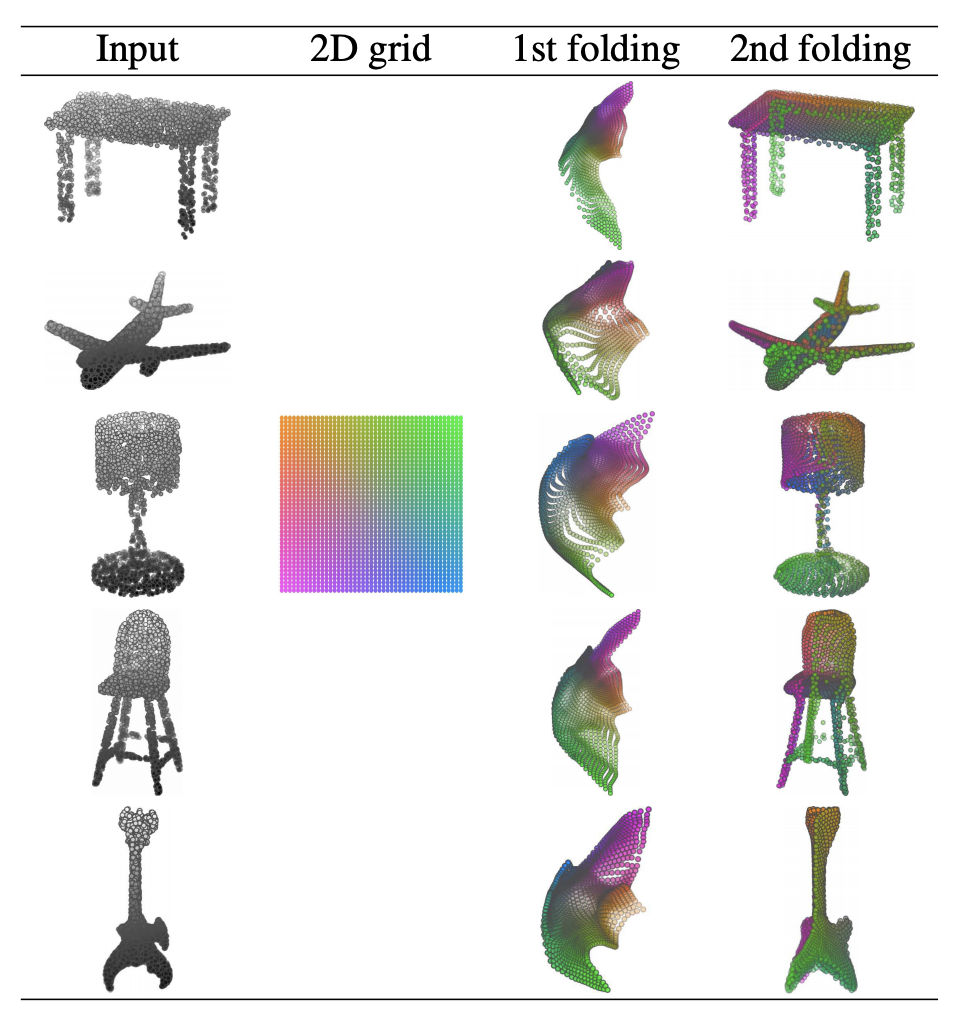}
    \caption{Description of two-step-folding decoding. The second column shows the 2D grid to be folded when decoding. The third column consists of the results underwent one folding operation. The fourth column composes of the results after the two folding. The final result is also the recovered point cloud. The color gradient is used to explain the correspondence between the 2D grids in the second column and the recovered point clouds in the last two ones after folding \cite{r55}.}
    \vspace{-0.3cm}
\end{figure}

Folding-based methods (KCNet \cite{r56}, MSN \cite{r22}, and PoinTr\cite{r89}) usually sample 2D grids from a 2D plane with fixed size and then concatenate them with the global shape representation extracted by the point cloud feature encoder. KCNet \cite{r56}, AtlasNet \cite{r12}, MSN\cite{r22} and SA-Net \cite{r58} reconstruct the complete object by evaluating a set of parametric surface elements and learn projections from 2D to 3D surface elements. SA-Net proposes a structure-preserving decoder with hierarchical folding for complete shape generation. 

\begin{figure}[ht]
    \centering
    \includegraphics[width=21pc]{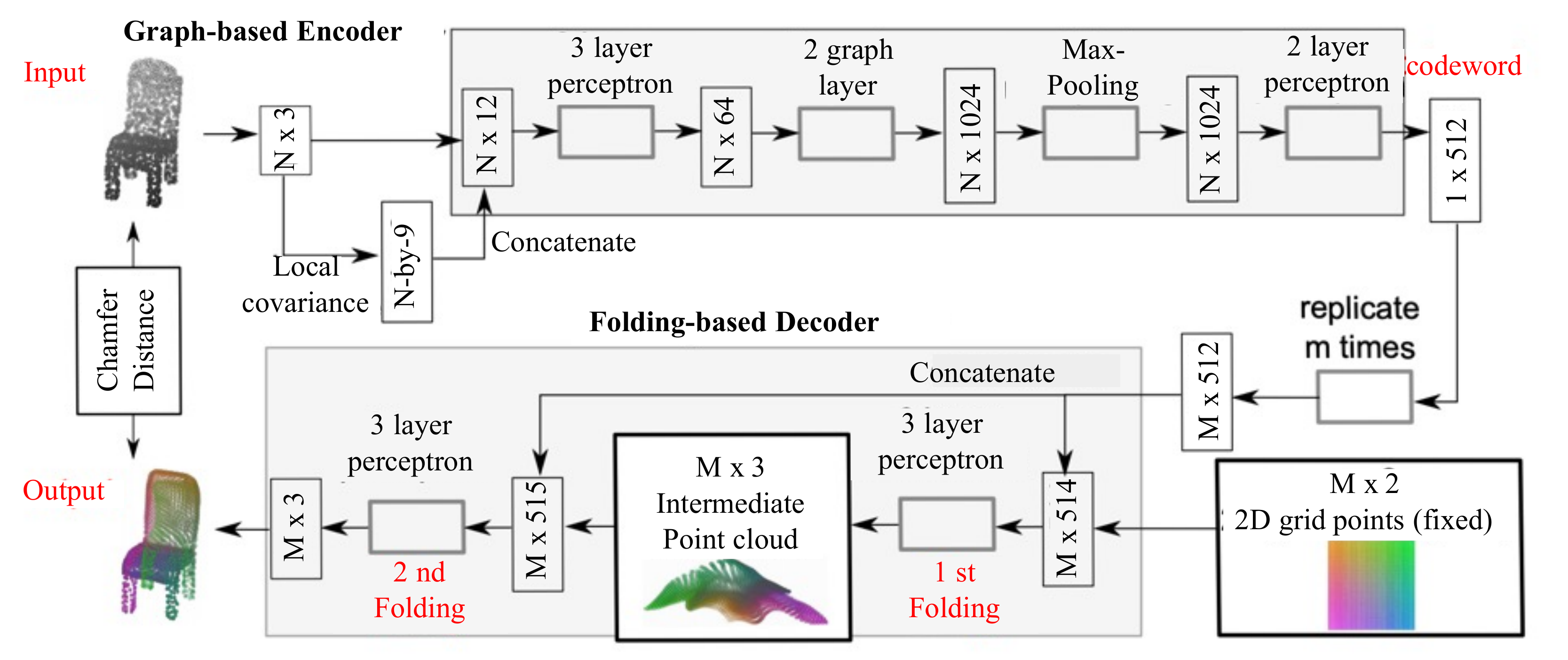}
    \caption{The architecture of FoldingNet \cite{r55}.}
    \vspace{-0.3cm}
\end{figure}

Moreover, TopNet \cite{r16} explores the hierarchical root tree architecture as a decoder to produce a random grouping of points and visually demonstrates the architecture exploited by the decoder through visualizing a node in the tree decoder as a collection of its children. To fully utilize structure details, Wen et al. \cite{r58} proposed Skip-Attention Network, which contributed two aspects: A skip-attention mechanism was adopted to explore the regional structure details of partial inputs, and a structure-preserving decoder using hierarchical folding was proposed to make use of the selected geometric information.

Despite their limited success, the great details of an object are often missed. Existing folding-derived methods, such as PCN \cite{r17}, FoldingNet \cite{r55} and TopNet \cite{r16}, fail to produce the structure details of an object to certain extent. One of the reasons is that they only rely on a single global shape representation to predict the entire point cloud. In contrast, the rich local region information that helps recover the detailed geometry is not fully utilized. Zong et al. \cite{r59} proposed an adaptive sampling and hierarchical folding network (ASHF-Net), in which the denoising auto-encoder with an adaptive sampling module to learn local region features, while the hierarchical folding decoder with the gated skip-attention and multi-resolution completion objectives make use of local structural details. Li et al. \cite{r99} combined a point-based encoder with an FC-based decoder and a folding-based decoder to produce the complete output, and this model with multistage loss function can be directly applied on the completion of point clouds.

Up-to-now, FoldingNet is the most widely used decoding block in the existing point cloud completion network. There is a drawback in FoldingNet, promoting researchers to build new decoder blocks. The folding manipulation samples the same 2D grids for each parent point, overlooking the local shape characteristics contained in the parent points.

However, there are some limitations of point-based methods.
\begin{itemize}
    \item The point-based network mainly tackles the permutation issue. Although the point-based methods treat points independently at the local level to maintain permutation invariance, this independence overlooks the geometric relations between the points and their neighbors. It has a fundamental limitation, leading to the loss of local features.
    \item Most point-based methods work in a coarse-to-fine manner. They are struggling to reconstruct object details, mainly due to two reasons: 1) the coarse outputs created from global embeddings lose the high-frequency information for 3D shapes; 2) the second stage acts as a point up-sampling function that fails to synthesize complex topologies.
    \item The point-based models deal directly with the points and have an extensive computation, which is inferior to the voxel-based method in large scenarios.
\end{itemize}

\subsection{View-based methods} By the merits of the modality of the images, the key challenge to solve the completion of the point cloud is to effectively integrate the features brought by pose and regional details derived from the incomplete and the global shape information from the single-view images (Fig. 8). As a sensor fusion network, Zhang et al. \cite{r44} proposed ViPC, which is a view-guided architecture. The ViPC retrieves the missing global structural information from an additional single-view image. The main contribution of ViPC lies in "Dynamic Offset Predictor," which could refine the coarse output. A multi-view consistent inference was proposed by Zwicker et al. \cite{r36} to strengthen geometric consistency in view-based 3D shape completion. And a multi-view consistency loss algorithm for inference optimization was defined, which could be implemented without ground truth supervision. Besides, the depth scan is utilized in ME-PCN \cite{r118} to make networks sensitive to shape boundaries, which enables ME-PCN to recover fine-grained surface details while keeping consistent local topology. In order to estimate the 6-DoF pose of 3D canonical shape with the help of several partial observations derived from the same object, Gu et al. \cite{r68} proposed a weakly-supervised approach to solve this issue. In the training process, the network uses multi-view geometric constraints to jointly optimize the canonical shapes and poses, which can deduce the complete results under the condition of a single partial input. A multi-view completion net (MVCN) \cite{r69}, combining GAN and multi-view information to enhance the performance of point cloud completion.

The main character and drawback can be listed as follow:
\begin{itemize}
    \item Different from other methods, the inputs of view-based methods are images, which might be RGB-D images or depth images.
    \item The performance will largely be depended on the angle and number of views due to the different information that can be acquired from these images.
\end{itemize}

\subsection{Convolution-based methods}

Encouraged by the great success of convolutional neural networks (CNNs) on 2D images, several works try to utilize 3D CNNs to learn the volume representation of three-dimensional point clouds. Nevertheless, transforming a point cloud into 3D volume will bring quantization effects: (1) Loss of details; (2) Insufficient to represent fine-grained information. Therefore, as far as we know, some works directly apply CNN on irregular, partial, and defect point clouds for 3D shape completion.

\begin{figure}[ht]
    \centering
    \includegraphics[width=21pc]{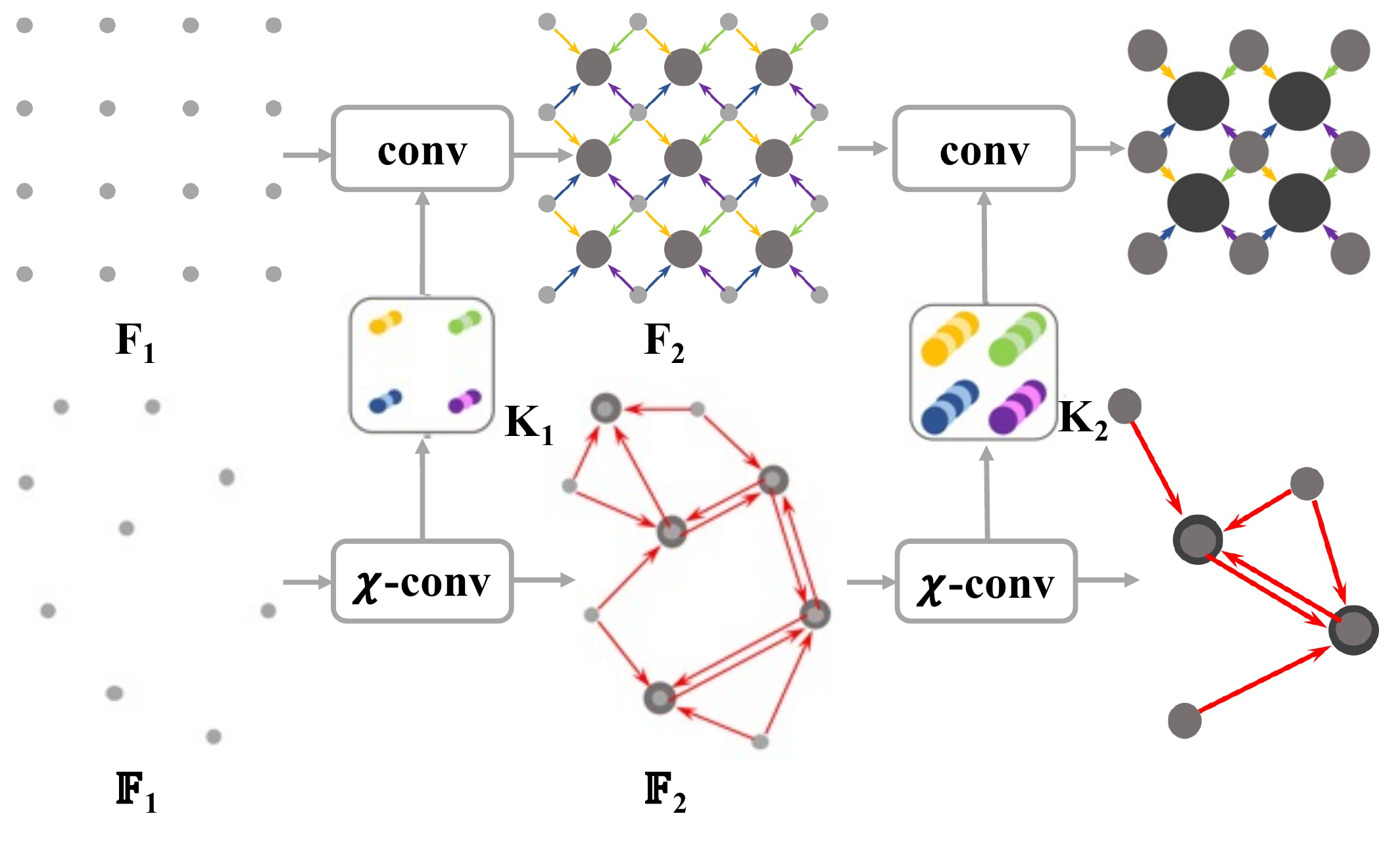}
    \caption{The illustration of hierarchical convolution on regular grids (top) and point clouds (bottom). In regular grids, convolutions are recursively performed on local grid patches, which typically reduces the grid resolution $(4 \times 4 \rightarrow 3 \times 3 \rightarrow 2 \times 2)$, while increasing the number of channels. In point cloud, $\mathcal{X}$-Conv is recursively applied to project or aggregate information from neighborhoods into fewer representative points but with richer information. $(9 \rightarrow 5 \rightarrow 2)$\cite{r38}.}
    \vspace{-0.3cm}
\end{figure}

\textbf{Preliminary works} In terms of the processing of point clouds, several contributions developed CNNs acting on discrete 3D grids of point cloud transformation. Hua et al. \cite{r37} defined convolution kernels on regular 3D grids, where the points are given the same weights when falling into the same grid. PointCNN \cite{r38} implements permutation invariance through a $\mathcal{X}$-conv transformation. In addition to CNN on discrete spaces, several methods define convolution kernels on continuous space (Fig. 11). A rigid and deformable kernel convolution (KPConv) module was devised by Thomas et al. \cite{r39} to utilize a collection of learnable kernel points for 3D point clouds. The dynamic filter was extended into a convolution operator dubbed PointConv by Tao et al. \cite{r40}. This operator could be employed to fulfill the deep convolution architecture. 

\textbf{Convolutional encoder} In this field, the point cloud will first be voxelized as input of 3D CNNs. Implicit Feature Network (IF-Nets) was devised by Pons-Moll et al. \cite{r42} to deliver continuous outputs that can handle multiple topologies and complete shapes for incomplete or sparse input data. Still, critically they can also retain details when it is present in the input data and can reconstruct articulated humans. Funkhouser et al. \cite{r43} devised the Sparse Voxel Completion Network (SVCN), which is composed of two U-Net-like sub-net for structure generation and refinement, respectively. The structure generation sub-net converts the input data into a set of sparse voxels by voxelization and outputting denser voxels representing the 3D surfaces. Then redundant voxels are deleted from the structure refinement network.

However, the voxelization process leads to irreversible loss of geometric information. Xie et al. \cite{r41} introduced a Gridding Residual Network (GRNet) and took the 3D grids as intermediate representations to process irregular point clouds. In GRNet, Gridding and Gridding Reverse methods were designed to transform point clouds into 3D grids without any loss of structural information. And the Cubic Feature Sampling layer was presented to extract information of adjacent points and preserve context knowledge. GRNet enables the convolutions on 3D point clouds while preserving their structural and context information. However, the voxel representation of GRNet is only used to reconstruct low-resolution shapes. Therefore, Wang et al. \cite{r119} develop VE-PCN to transform the unordered point sets into grid representations to support edge generation and point cloud reconstruction. This multi-scale VE-PCN is able to generate fine-grained details for point cloud completion. Liu et al. \cite{r120} presented MRAC-Net, which includes an anisotropic convolutional encoder for extracting local and global features to enhance the model’s extraction ability of latent features.

\textbf{Deconvolutional decoder} Except for feature learning, convolutions can also be utilized in reconstructing the point clouds. Wang et al. \cite{r100} designed SoftPoolNet, which organizes the extracted features by PointNet called soft pool according to their activation. Regional convolutions were designed to maximize the global activation entropy for the decoding stage. To recover the details of the point clouds and retain the original plane structures, Deng et al. \cite{r123} proposed 3D Grid Transformation Network, where the weights were calculated for the reconstructed point clouds.

Unlike PointCNN, KPConv and PointConv, in the task of point cloud completion, nearly all convolution-based methods tend to voxelize the point cloud before applying 3D convolution. Hence, we mainly discuss the limitations of these convolution-based methods in volumetric 3D data representation:
\begin{itemize}
    \item First, not all voxels or grid representations are helpful because they contain occupied and non-occupied parts of the scanning environments. Thus, the high demand for computer storage is unnecessary within this ineffective data representations.
    \item Second, the voxel or grid size is hard to set, affecting the scale of input data and may disrupt the spatial relationship between points.
    \item Third, computation and memory requirements grow cubically with the resolution.
\end{itemize}

\subsection{Graph-based methods}

Since both point clouds and graphs can be regarded as non-Euclidean structured data, exploring the relationship between points or local regions by taking them as the vertices of some graphs is convenient (Fig. 12). Regarding every point in the inputs as the vertices, the edges could be generated by graph-based networks based on the adjacent points. Hence, graph convolutions are naturally suitable for the processing of point clouds. These methods use the advantages of graph convolutions, usually convolving the spatial neighborhoods and generating a new graph by gathering the neighborhood information of each point. Compared to point-based methods, the graph-based approaches consider regional geometric details. 

\begin{figure}[ht]
    \centering
    \includegraphics[width=21pc]{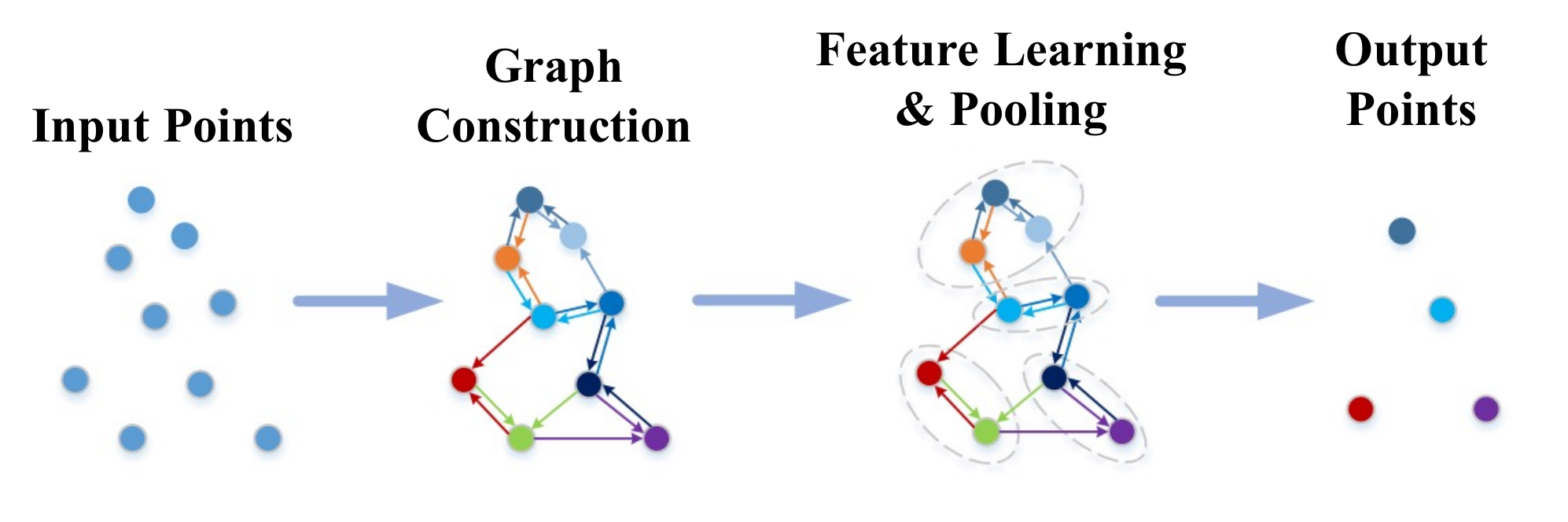}
    \caption{An illustration of a graph-based network \cite{r2}.}
    \vspace{-0.3cm}
\end{figure}

As a pioneering work, a dynamic graph convolution was introduced by DGCNN \cite{r45}. In the dynamic graph convolution, adjacent matrices could be calculated by the relations of vertexes, which come from latent space. The graph is established in the feature space and can be dynamically updated in the DGCNN. Moreover, EdgeConv was devised to calculate graphs in every network layer dynamically and could be integrated with the existing architectures(Fig. 13). In addition, LDGCNN \cite{r46} removes the transformation and connects the multi-level features learned in different layers in DGCNN. Thus, the performance and model size can be optimized. Stimulated by DGCNN, Hassani, and Haley \cite{r47} introduced the multi-level network to exploit points and shape features for self-supervised reconstruction. Furthermore, following DGCNN, DCG \cite{r116} encodes regional links as feature vectors and refines the point clouds in a coarse-to-fine manner.

\begin{figure}[ht]
    \centering
    \includegraphics[width=21pc]{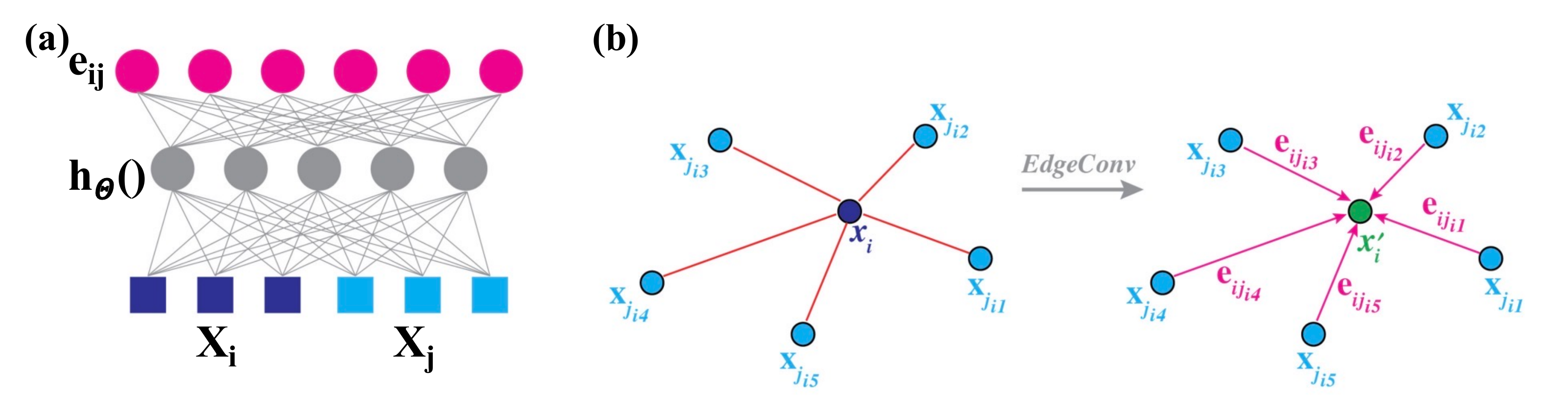}
    \caption{(a) An edge feature $e_{ij}$ is calculated from the pair of point $x_i$ and $x_j$. (b) The EdgeConv operator. The output of EdgeConv is computed by converging the edge features of all edges emitted by each connected vertex\cite{r45}.}
    \vspace{-0.3cm}
\end{figure}

Except for the dynamic graph convolution, PointNet++ \cite{r11} and FoldingNet \cite{r55} could also be regarded as a type of method using graph convolution to exploit information from fixed adjacencies of sampled center points. Combining with graph convolution, Pan \cite{r48} designed a hierarchical encoder to refine the local geometric details by propagating multi-scale edge features, which were captured by skeleton generation. Following that, the Edge-aware Feature Expansion (EFE) module was proposed to expand/up-sample information of points by highlighting the regional edge of the point. The ECG can preserve both global structural information and local pattern features. Nodeshuffle and Inception DenseGCN \cite{r49} were proposed by Qian et al. The former utilizes a Graph Convolutional Network (GCN) to encode regional point features from adjacent points better, while the latter aggregates features at multiple scales. The PU-GCN is a new point up-sampling pipeline when combining the Inception DenseGCN with NodeShuffle \cite{r77}. Shen et al. \cite{r50} proposed a Graph Guided Deformation Network, where the input and intermediate data were considered as controlling and supporting points, respectively, and modeled the optimization guided by the graph convolutional network for the task of point cloud completion. This network simulates the least square Laplace deformation process via mesh deformation methods, which has the adaptive ability to model the geometric details of the modeling and reduce the gap between the mesh deformation algorithm and the point cloud completion task. Li et al. \cite{r51} designed the PRSCN, which firstly uses Point Rank Sampling approaches to rate and sample feature points more objectively through local outline form. Subsequently, considering the connections among features from different scales, a Cross-Cascade block was designed to integrate features. Besides, Leap-type EdgeConv was integrated to expand the receptive field with kernel size maintained. Moreover, by utilizing global features and local features, LRA-Net \cite{r98} was proposed to recover complete point clouds with more details and smoother shapes, which are derived from the structure of PointNet and Graph Convolutional Network (GCN).

\textbf{Attention-assisted GCN} Furthermore, the attentional mechanism is also introduced into GCN. To recover fine-grained shapes, Wu et al. \cite{r52} introduced a learning-based method. They sample local regions of partial inputs, encode their features, and combine them with exploited global features. After the graph is built, all the regional features are gathered, and the graph is convolved with multi-head attention. Graph attention mechanism enables each local feature vector to be searched across the regions and selectively absorb other local features based on relationships in high-dimensional feature space. CRA-Net \cite{r53} designed a cross-regional attention unit based on graph attention. This module quantifies underlying connections among regional features under specific contexts and is explained by global features. Given such links, every conditional regional feature vector can be searched as graph attention. In PC-RGNN \cite{r54}, a graph neural network module was designed, which captures the relations between points comprehensively through the local-global attention mechanism and context aggregation based on a multi-scale graph, greatly enhancing the learning of features.

But there are two challenges for constructing graph-based networks as follows:
\begin{itemize}
    \item First, defining an operator suitable for dynamically sized neighborhoods and maintaining the weight sharing scheme of CNNs.
    \item Second, exploiting the spatial and geometric relationships among each node’s neighbors.
\end{itemize}

\subsection{GAN-based methods}

Compared to traditional CNN, the GAN \cite{r60} architecture utilizes a discriminator implicitly learning to estimate the point collections provided by the generator (Fig. 14). Due to the characteristics of 3D data, the integration of GAN in point cloud completion possesses several inherent challenges: 
\begin{itemize}
    \item Different from the grid structures of 2D images, where the locations of pixels are clearly defined. In contrast, point clouds with different 3D shapes are highly unstructured. In general, GANs trained on 3D shapes produce point clouds with significant inhomogeneity. That is, points are not evenly distributed on the shape's surface. This inhomogeneity can lead to shapes with unwanted holes, undermining the integrity of predictions.
    \item The disorder of point clouds determines the completion task significantly different from two-dimensional image completion. In 2D image rendering, one can easily measure the reconstructed consistency between partial input visible regions and predicted outputs given mesh aligned pixel corresponding. This comparison is challenging in 3D shape completion because the corresponding regions of two 3D shapes may be located at different positions in 3D space. GAN inversion results in poor reconstruction, which jeopardizes the shape-completion mission.
    \item Whereas simple GANs can only yield a small size of (1024 or 2048) point collections because of the complicated point distribution and the notoriously hard training of GANs. 
\end{itemize}

Therefore, the researchers greatly improved the point cloud completion based on the traditional GAN.

\begin{figure}[ht]
    \centering
    \includegraphics[width=10pc]{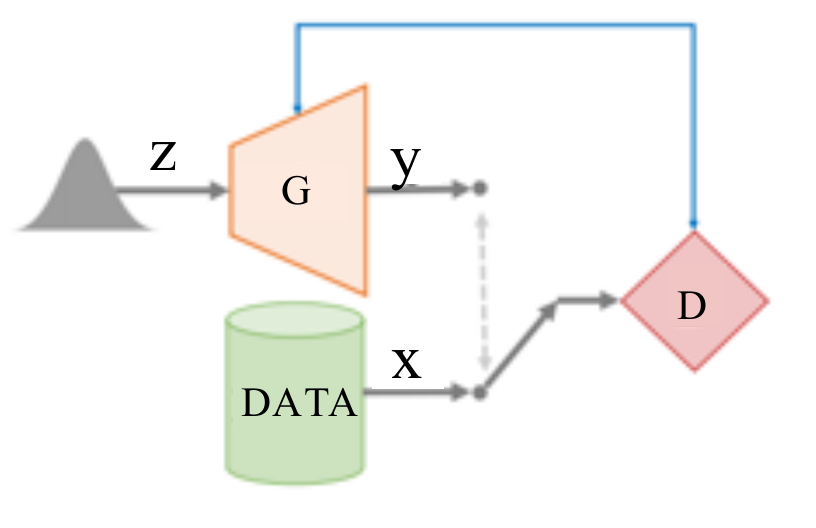}
    \caption{An description of a generative adversarial networks. The basic architecture is on basis of antagonism between a generator ($G$) and a discriminator ($D$). The $G$ is designed to produce points that seem different from real data ($x \sim p_{data}$) via a random sample from a simple distribution $z \sim p_z$ through the generator function. The task of the discriminator is to distinguish synthetic samples from real ones.}
    \vspace{-0.3cm}
\end{figure}

\textbf{End-to-end mechanism} End-to-end learning is commonly used in 3D point cloud completion. By the integration of a 3D Encoder-Decoder Generative Adversarial Network (3D-ED-GAN) and a Long-term Recurrent Convolutional Network (LRCN), Wang et al. \cite{r61} introduced a new structure. 3D-ED-GAN leverages an encoder to map voxelized 3D shapes into a probabilistic latent space and uses a GAN to facilitate the decoder generating the complete volumetric shapes with the help of the latent feature representations. Nevertheless, these approaches can only use 3D volumes as inputs or get a voxel representation for results. Achlioptas et al. \cite{r62} devised r-GAN with both generator and discriminator using fully connected layers. The AE was trained to learn the latent space. The l-GANs were trained in the latent space, which was more easily trained than simple GANs with coverage of data distributions. In the training of latent representation, the capability of multi-class GAN is almost the same as that of class-specific GAN. Gurumurthy et al. \cite{r63} have devised a scheme utilizing latent GAN and AE. Nevertheless, they used a time-consuming optimization procedure for every batch of inputs to choose the best seed for GAN. Yu et al. \cite{r65} devised a point encoder GAN, where the max-pooling layer was utilized to tackle the irregular problem in the learning process, and two T-Nets (derived from PointNet) were added in the encoder-decoder architecture to represent the characteristics of the inputs better. And a hybrid recovery loss function was proposed to compute the diversity between two groups of disordered data. Chen et al. \cite{r66} proposed an end-to-end conditional GAN called GeneCGAN. From the heredity perspective, a simulated genetic (SG) layer was devised. It is executed in a hierarchical root tree using ancestor information and the connection of neighborhoods. Through a prior fusion strategy, the global feature is appended to the tree's root node as conditional information, and the conditional probability distribution of the inputs is learned.

% Achlioptas et al. \cite{r64} introduced a deep AE network with great recovery performance and generalization capacity. In the 3D recognitions, the shape transformation is realized by algebraic operation. They conducted in-depth studies on different generative models, such as GAN operated on the original inputs, GANs significantly trained in the AEs fixed latent space, together with Gaussian mixture models (GMMS).

\begin{figure}[ht]
    \centering
    \includegraphics[width=21pc]{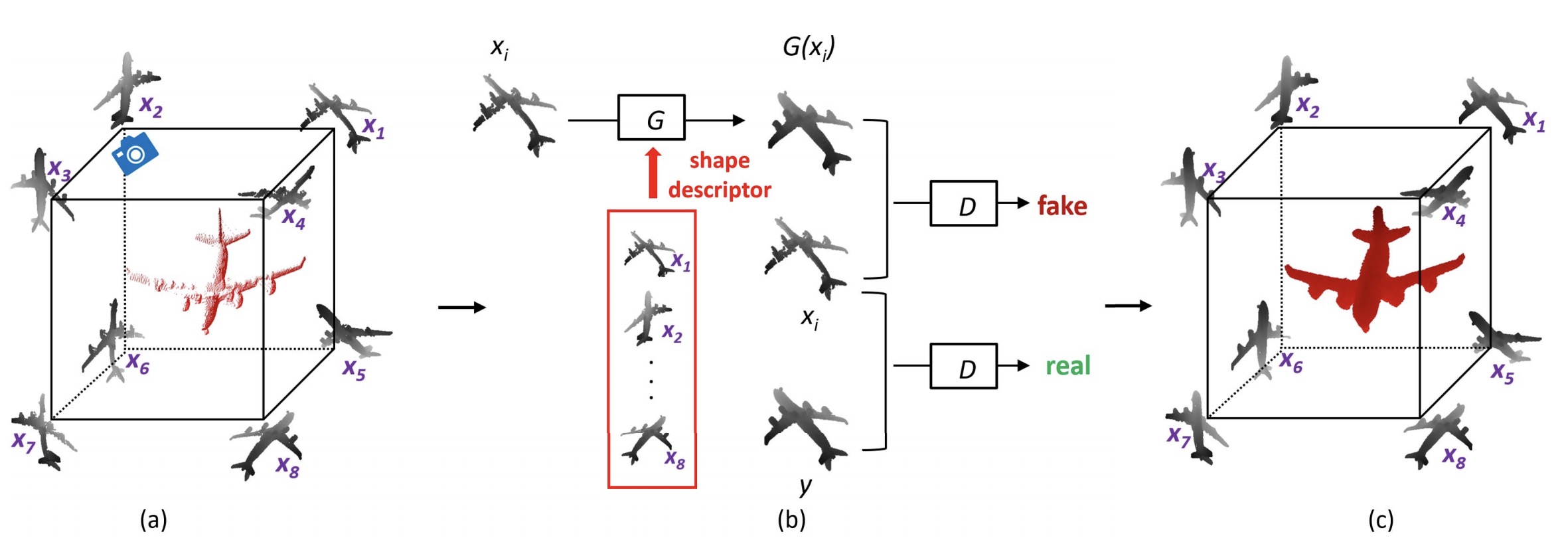}
    \caption{An illustration of a generative multi-view representation. For example, (a) 8 depth maps of defect inputs were rendered from 8 viewpoints; (b) These introduced 8 depth maps are completed through a multi-view completion network, including adversarial loss, producing 8 completed depth maps; (c) The 8 depth maps are back-projected into a completed shape.\cite{r69}}
    \vspace{-0.3cm}
\end{figure}

The effective latent space representations of the point cloud provide important and fundamental information that can be utilized for 3D shape reconstruction. Wen et al. \cite{r73} presented Cycle4Completion with two synchronous cycle transformations between the latent spaces of complete 3D shapes and incomplete 3D shapes. The cycle transformation could facilitate the model to learn about 3D shapes through studying to produce complete or incomplete shapes with the help of complementary shapes. Chen et al. \cite{r74} proposed a framework for unpaired shape completion, the core of which is an adaptation network that acts as a generator to transform latent codes of the original point scans and maps them to underlying latent spaces of clean and complete object scans. These two latent spaces regularize the issue by limiting the transfer issue to respective data manifolds. Zhang et al. \cite{r75} presented ShapeInversion to introduce GAN inversion into shape completion for the first time. ShapeInversion leverages a GAN pre-trained on complete shapes by searching for a latent code to obtain the complete shape that best reconstructs the input of a given part. In this way, ShapeInversion eliminates the need for paired training data and can combine rich prior information in a well-trained generation model. Combining the latent-space GAN and Laplacian GAN, Egiazarian et al. \cite{r76} devised a multi-level network that can produce 3D objects with increasing levels of details. Based on GAN, a PU-GAN was presented by Li et al. \cite{r77}, which is an up-sampling network that aims to learn the point distributions from the latent space. The PU-GAN can also up-sample points over patches on the surfaces of the object. An up-down-up expansion module was constructed in the generator to up-sample point features, which has an error feedback unit and self-correction functions. Moreover, a self-attention unit was also developed to increase the integration of features.

Furthermore, Li et al. \cite{r80} proposed a two-fold modification of GAN (PC-GAN). In PC-GAN, hierarchical Bayesian networks and implicit generative architectures are combined through hierarchical and interpretable sampling. The key of this approach lies in the posterior inference model, which is trained for hidden variables. Besides, rather than using the SOTA Wasserstein GAN objective, a sandwiching objective was devised to result in a compacter Wasserstein distance estimate than the typically utilized dual form. Hence, PC-GAN provided a generic architecture to comprise the existing GAN easily. Tao et al. \cite{r81} devised a dual-generators network, in which the first generator was designed to learn point embeddings, while the second one was utilized to refine the generated point cloud based on a depth-first point embedding to generate a uniform output. To minimize the influence of noises and geometrical loss of incomplete point cloud, PF-Net \cite{r82} retains the spatial arrangement of incomplete inputs and can calculate the complicated geometry of the missing area. To this end, PF-Net uses a multi-level generation based on feature points to predict the missing parts in a hierarchical network. PF-Net also utilizes multi-stage completion and adversarial loss to produce more realistic missing regions. Among them, the adversarial loss can better solve the problem of multiple modes in the prediction. Cheng et al. \cite{r121} proposed an end-to-end generative adversarial network-based dense point cloud completion architecture (DPCG-Net). The DPCG-Net designed two GAN-based modules that translate point cloud completion into mapping between global feature distributions obtained by encoding partial point clouds and ground truth, respectively.

\textbf{Refinement} Besides, the refinement strategy is also commonly integrated with GAN. Wang et al. \cite{r70} developed a feature alignment approach for learning the shape prior. Moreover, a coarse-to-fine method was devised to combine the shape prior with the fine phase. The loss of feature alignment comprises an L2 distance and an adversarial loss derived from Maximum Mean Discrepancy Generative Adversarial Network (MMD-GAN). Wang et al. \cite{r71} devised a point completion network with a cascaded refinement network (CRN) as a generator to synthesize these missing parts with high quality by using the details of inputs. Moreover, they designed a patch discriminator, which uses adversarial training to know the precise point distribution and punishes the generated shapes differently from the ground truth. Furthermore, to generate high-quality targets with detailed geometry, Wang et al. \cite{r72} expanded this strategy to synthesize fine-grained targets. Two self-training strategies were proposed to improve the reconstruction performances in both supervised and self-supervised settings.

\textbf{Multi-view GAN} The views of the same 3D model share some common information that can be explored, including global and regional information as seen from various angles. Zwicker et al. \cite{r69} proposed a multi-view completion net (MVCN) (Fig. 15), which leverages information from all views of a 3D shape to assist the completion of every single view. Benefiting from the multi-view presentations and network architecture with conditional GAN, MVCN enhances the performance of 3D completion. Liu et al. \cite{r67} tried to convert the three-dimensional point cloud generation issue to a two-dimensional geometry-image generation problem and introduced an adversarial VAE to optimize the GIG proposed by combining adversarial learning with VAE. While it is easy to create depth maps of 3D shapes independently, there are two drawbacks. First, they do not encourage consistency between depth maps from the same 3D object, influencing the accuracy of 3D objects obtained through back-projecting completed depth maps. Second, they could not complete a depth map and use information from other depth maps of the same 3D object. The accuracy of completing a single depth map is limited.

\textbf{Integrated with RL} More recently, reinforcement learning (RL) has been integrated into GAN. Sarmad et al. \cite{r78} presented RL-GAN-Net, in which a reinforcement learning mechanism can control the GAN. The architecture can transform noisy defect input data into full shape with high fidelity through the control of GAN. Vaccine-Style-Net \cite{r79} was carried out in the function space of 3D surfaces, and the 3D surface is represented as the continuous decision boundary function. At the same time, an RL unit is embedded to derive the complete 3D geometry from the partial inputs.

\textbf{Integrated with GCN} Except for RL, GCN is also commonly utilized with GAN. Valsesia et al. \cite{r94, r95} studied the unsupervised problem of generative models using graph convolution. They emphasize the generator of GAN and define graph convolution methods. They devised a network that learns to produce local features to approximate the embedding of output geometry and defined an up-sampling layer of a graph convolution to sample more efficiently using self-similarity prior. Xie et al. \cite{r96} presented a SpareNet by utilizing channel-attentive EdgeConv to learn the regional features and global shape. SpareNet utilizes shape features as style code to adjust the normalization layer during folding to enhance its capabilities. Moreover, a differentiable renderer is used to project the complete point cloud onto the depth map, and adversarial training is applied to promote the perception of reality from different viewpoints. Li et al. \cite{r97} devised a Hierarchical Self-Attention GAN (HSGAN) to use a random code and transforms it hierarchically into a representation graph by combining GCN and self-attention. In this model, the topology of the global graph is embedded into shape generation, and latent topology information is utilized to recover the geometry structure of the 3D shape.

% Please add the following required packages to your document preamble:
% \usepackage{multirow}
% \usepackage{graphicx}
\begin{table*}[htbp]\footnotesize

\caption{Comparison of 3D point clouds completion performances on the PCN. Here, 'CD' represents the mean Chamfer Distance, and 'EMD' represents the mean Earth Mover’s Distance. The '-' represents the performances that are unreachable. (The CD loss is scaled by 1000, and EMD loss is scaled by 100. If not specified, the CD means CD-$\ell1$)}
\renewcommand{\arraystretch}{1.03}
\centering
{%
\begin{tabular}{|cc|ccc|}
\hline
\multicolumn{2}{|c|}{\multirow{1}{*}{Methods}}                                                                                                                                                 & \multicolumn{3}{|c|}{\multirow{1}{*}{PCN}}                                                                                                                                                                         \\ \cline{3-5} 
\multicolumn{2}{|c|}{}                                                                                                                                                                         & \multicolumn{1}{c|}{\begin{tabular}[c]{@{}c@{}}CD \\ (Value/Resolution)\end{tabular}}                                              & \multicolumn{1}{c|}{\begin{tabular}[c]{@{}c@{}}EMD \\ (Value/Resolution)\end{tabular}}   & \begin{tabular}[c]{@{}c@{}}F-Score@1\% \\ (Value/Resolution)\end{tabular} \\ \hline
\multicolumn{1}{|c|}{\multirow{1}{*}{\begin{tabular}[c]{@{}c@{}}Point-based \\  Methods\end{tabular}}}      & N-DPC\cite{r26}                                                                            & \multicolumn{1}{c|}{10.25/16384}                                                                                                   & \multicolumn{1}{c|}{6.05/16384}                                                          & -                                                                         \\ \cline{2-5} 
\multicolumn{1}{|c|}{}                                                                                      & MSN \cite{r22}                                                                              & \multicolumn{1}{c|}{10.00/8192}                                                                                                       & \multicolumn{1}{c|}{3.78/8192}                                                           & -                                                                         \\ \cline{2-5} 
\multicolumn{1}{|c|}{}                                                                                      & MSPCN w/ MVCS \cite{r25}                                                                    & \multicolumn{1}{c|}{0.94/16384 (CD-$\ell_2$)}                                                                                                    & \multicolumn{1}{c|}{-}                                                                   & -                                                                         \\ \cline{2-5} 
\multicolumn{1}{|c|}{}                                                                                      & PCN \cite{r17}                                                                             & \multicolumn{1}{c|}{10.02/16384}                                                                                                     & \multicolumn{1}{c|}{6.40/16384}                                                          & -                                                                         \\ \cline{2-5} 
\multicolumn{1}{|c|}{}                                                                                      & SK-PCN\cite{r24}                                                                           & \multicolumn{1}{c|}{0.18/2048 (CD-$\ell_2$)}                                                                                                    & \multicolumn{1}{c|}{0.50/2048}                                                          & -                                                                         \\ \cline{2-5} 
\multicolumn{1}{|c|}{}                                                                                      & ASFM-Net\cite{r31}                                                                         & \multicolumn{1}{c|}{12.09/4096}                                                                                                    & \multicolumn{1}{c|}{-}                                                                   & -                                                                         \\ \cline{2-5} 
\multicolumn{1}{|c|}{}                                                                                      & FinerPCN\cite{r23}                                                                         & \multicolumn{1}{c|}{16.65/2048}                                                                                                    & \multicolumn{1}{c|}{6.14/2048}                                                           &                                                                           \\ \cline{2-5} 
\multicolumn{1}{|c|}{}                                                                                      & \begin{tabular}[c]{@{}c@{}}Refinement of Predicted \\ Missing Parts\cite{r23}\end{tabular} & \multicolumn{1}{c|}{1.28/2048}                                                                                                   & \multicolumn{1}{c|}{-}                                                                   & -                                                                         \\ \cline{2-5}

\multicolumn{1}{|c|}{}                                                                                      & SDME-Net\cite{r34}                                                                         & \multicolumn{1}{c|}{17.00/4096}                                                                                                   & \multicolumn{1}{c|}{4.33/4096}                                                           & -                                                                         \\\cline{2-5} 
\multicolumn{1}{|c|}{}                                                                        & ASHF-Net\cite{r59}                                                                         & \multicolumn{1}{c|}{0.26/16384}                                                                                                  & \multicolumn{1}{c|}{-}                                                                   & -                                                                         \\ \cline{2-5} 
\multicolumn{1}{|c|}{}                                                                                      & TopNet\cite{r16}                                                                            & \multicolumn{1}{c|}{0.97/2048}                                                                                                    & \multicolumn{1}{c|}{-}                                                                   & -                                                                         \\ \cline{2-5} 
\multicolumn{1}{|c|}{}                                                                                      & SA-Net\cite{r58}                                                                           & \multicolumn{1}{c|}{0.77/2048}                                                                                                    & \multicolumn{1}{c|}{-}                                                                   & -                                                                         \\ \cline{2-5} 
\multicolumn{1}{|c|}{}                                                                                      & \multicolumn{1}{l|}{Multistage Loss Function\cite{r99} }                                    & \multicolumn{1}{c|}{8.75/1024}                                                                                                    & \multicolumn{1}{c|}{5.02/1024}                                                          & -                                                                         \\ \hline
\multicolumn{1}{|c|}{\multirow{1}{*}{\begin{tabular}[c]{@{}c@{}}View-based\\  Methods\end{tabular}}}  &
\begin{tabular}[c]{@{}c@{}}Multi-View \\ Consistent Inference \cite{r36}\end{tabular}       & \multicolumn{1}{c|}{8.05/2048}                                                                                                    & \multicolumn{1}{c|}{-}                                                                   & -                                                                         \\ \cline{2-5} \multicolumn{1}{|c|}{}                                                                                      & 
\begin{tabular}[c]{@{}c@{}}Weakly-Supervised \\ 3D Shape Completion\cite{r68}\end{tabular} & \multicolumn{1}{c|}{\begin{tabular}[c]{@{}c@{}}26.40/8196 \\ (3 categories)\end{tabular}}                                           & \multicolumn{1}{c|}{-}                                                                   & -                                                                         \\ \cline{2-5} 
\multicolumn{1}{|c|}{}                                                                                      & \begin{tabular}[c]{@{}c@{}}MVCN \\ (Render4Completion)\cite{r69} \end{tabular}              & \multicolumn{1}{c|}{8.30/16384}                                                                                                    & \multicolumn{1}{c|}{-}                                                                   & -                                                                         \\ \cline{2-5} \multicolumn{1}{|c|}{}                                                                                      & ViPC\cite{r44}                                                                             & \multicolumn{1}{c|}{\begin{tabular}[c]{@{}c@{}}3.31/2048 \\ (ShapeNet-ViPC)\end{tabular}}                                   & \multicolumn{1}{c|}{-}                                                                   & \begin{tabular}[c]{@{}c@{}}0.59/2048 \\ (ShapeNet-ViPC)\end{tabular}                                          \\ \hline
\multicolumn{1}{|c|}{\multirow{1}{*}{\begin{tabular}[c]{@{}c@{}}Convolution-based\\ Methods\end{tabular}}}  & GRNet\cite{r41}                                                                            & \multicolumn{1}{c|}{0.27/16384 (CD-$\ell_2$)}                                                                                                  & \multicolumn{1}{c|}{-}                                                                   & 0.71/16384                                                               \\ \cline{2-5} 
 
\multicolumn{1}{|c|}{}                                                                                      & SoftPoolNet\cite{r100}                                                                      & \multicolumn{1}{c|}{5.94/16384}                                                                                                    & \multicolumn{1}{c|}{4.80/1024}                                                           & -                                                                         \\ \hline
\multicolumn{1}{|c|}{\multirow{1}{*}{\begin{tabular}[c]{@{}c@{}}Graph-based \\  Methods\end{tabular}}}                                                                                    & ECG\cite{r48}                                                                              & \multicolumn{1}{c|}{1.02/2048}                                                                                                   & \multicolumn{1}{c|}{-}                                                                   & -                                                                         \\ \cline{2-5} 
\multicolumn{1}{|c|}{}                                                                                      & \begin{tabular}[c]{@{}c@{}}Graph-Guided \\ Deformation Network\cite{r50}\end{tabular}      & \multicolumn{1}{c|}{0.60/2048}                                                                                                    & \multicolumn{1}{c|}{-}                                                                   & -                                                                         \\ \cline{2-5} 
\multicolumn{1}{|c|}{}                                                                                      & PRSCN\cite{r51}                                                                            & \multicolumn{1}{c|}{\begin{tabular}[c]{@{}c@{}}4.48/2048 \\ (ShapeNet-part(14))\\ 4.78/2048 \\ (ShapeNet-part(16))\end{tabular}} & \multicolumn{1}{c|}{-}                                                                   & -                                                                         \\ \cline{2-5} 
\multicolumn{1}{|c|}{}                                                                                      & DCG\cite{r116}                                                                              & \multicolumn{1}{c|}{3.28/2048}                                                                                                    & \multicolumn{1}{c|}{-}                                                                   & -                                                                         \\ \cline{2-5} 
\multicolumn{1}{|c|}{}                                                                                      & LRA-Net\cite{r98}                                                                           & \multicolumn{1}{c|}{26.04/2048}                                                                                                   & \multicolumn{1}{c|}{11.21/2048}                                                         & -                                                                         \\ \hline
\multicolumn{1}{|c|}{\multirow{1}{*}{\begin{tabular}[c]{@{}c@{}}GAN-based\\  Methods\end{tabular}}}        & \begin{tabular}[c]{@{}c@{}}Graph Guided \\ Deformation\cite{r50} \end{tabular}              & \multicolumn{1}{c|}{0.60/2048}                                                                                                    & \multicolumn{1}{c|}{-}                                                                   & -                                                                         \\ \cline{2-5} 
\multicolumn{1}{|c|}{}                                                                                      & CRN\cite{r71}                                                                              & \multicolumn{1}{c|}{8.51/16384}                                                                                                    & \multicolumn{1}{c|}{-}                                                                   & -                                                                         \\ \cline{2-5} 
\multicolumn{1}{|c|}{}                                                                                      & \begin{tabular}[c]{@{}c@{}}Latent-Space \\ Laplacian Pyramids\cite{r76} \end{tabular}       & \multicolumn{1}{c|}{\begin{tabular}[c]{@{}c@{}}0.34/2048 \\ (3 categories)\end{tabular}}                                          & \multicolumn{1}{c|}{\begin{tabular}[c]{@{}c@{}}4.16/2048 \\ (3 categories)\end{tabular}} & -                                                                         \\ \cline{2-5} 
\multicolumn{1}{|c|}{}                                                                                      & \begin{tabular}[c]{@{}c@{}}MMD-GAN\cite{r70} \\ (Learning Shape Priors)\end{tabular}       & \multicolumn{1}{c|}{8.57/2048 (CD-$\ell_2$)}                                                                                                    & \multicolumn{1}{c|}{-}                                                                   & -                                                                         \\ \cline{2-5} 
\multicolumn{1}{|c|}{}                                                                                      & NSFA\cite{r35}                                                                        & \multicolumn{1}{c|}{0.81/16384}                                                                                                   & \multicolumn{1}{c|}{-}                                                                   & -                                                                         \\ \cline{2-5} 
\multicolumn{1}{|c|}{}                                                                                      & PF-Net\cite{r82}                                                                           & \multicolumn{1}{c|}{1.01/2048}                                                                                                    & \multicolumn{1}{c|}{-}                                                                   & -                                                                         \\ \cline{2-5} 
\multicolumn{1}{|c|}{}                                                                                      & \begin{tabular}[c]{@{}c@{}}SCRN\cite{r72}\end{tabular}           & \multicolumn{1}{c|}{8.51/16384}                                                                                                   & \multicolumn{1}{c|}{-}                                                                   & -                                                                         \\ \cline{2-5} 
\multicolumn{1}{|c|}{}                                                                                      & Cycle4Completion\cite{r73}                                                                  & \multicolumn{1}{c|}{1.41/2048}                                                                                                     & \multicolumn{1}{c|}{-}                                                                   & -                                                                         \\ \cline{2-5} 
\multicolumn{1}{|c|}{}                                                                                      & ShapeInversion\cite{r75}                                                                   & \multicolumn{1}{c|}{1.49/2048}                                                                                                     & \multicolumn{1}{c|}{-}                                                                   & 0.84/2048                                                                \\ \cline{2-5} 

\multicolumn{1}{|c|}{}                                                                                      & GencCGAN\cite{r66}                                                                          & \multicolumn{1}{c|}{2.63/2048}                                                                                                     & \multicolumn{1}{c|}{2.70/2048}                                                            & -                                                                         \\ \cline{2-5} 
\multicolumn{1}{|c|}{}                                                                                      & SpareNet\cite{r96}                                                                         & \multicolumn{1}{c|}{0.52/2048}                                                                                                    & \multicolumn{1}{c|}{0.19/2048}                                                         & -                                                                         \\ \hline
\multicolumn{1}{|c|}{\multirow{1}{*}{\begin{tabular}[c]{@{}c@{}}Transformer-based\\  Methods\end{tabular}}} & SnowflakeNet\cite{r90}                                                                     & \multicolumn{1}{c|}{1.86/2048}                                                                                                    & \multicolumn{1}{c|}{-}                                                                   & -                                                                         \\ \cline{2-5} 
\multicolumn{1}{|c|}{}                                                                                      & PoinTr\cite{r89}                                                                          & \multicolumn{1}{c|}{\begin{tabular}[c]{@{}c@{}}1.09/8192 (CD-$\ell_2$) \\ (ShapeNet-55)\end{tabular}}                                            & \multicolumn{1}{c|}{-}                                                                   &  {\begin{tabular}[c]{@{}c@{}}0.46/8192\\ (ShapeNet-55)\end{tabular}}                                                     \\ \hline
\multicolumn{1}{|c|}{Other Methods}                                                                         & PMP-Net\cite{r101}                                                                          & \multicolumn{1}{c|}{8.66/2048 }                                                                                                     & \multicolumn{1}{c|}{-}                                                                   & -                                                                         \\ \hline
\end{tabular}%
}
\vspace{-0.5cm}
\end{table*}

% Please add the following required packages to your document preamble:
% \usepackage{multirow}
% \usepackage{graphicx}
\begin{table*}[htbp]\footnotesize

\caption{Comparison of 3D point clouds completion performances on the ModelNet and Completion3D. 'CD' represents the mean Chamfer Distance and 'EMD' represents the mean Earth Mover’s Distance. The '-' stands for the performances are unreachable. (The CD loss is scaled by 1000 and EMD loss is scaled by 100.)}
\centering
\renewcommand{\arraystretch}{1.1}
{%
\begin{tabular}{|cc|cc|cc|}
\hline
\multicolumn{2}{|c|}{\multirow{1}{*}{Methods}}                                                                                                                                         & \multicolumn{2}{c|}{ModelNet}                                                                                                                             & \multicolumn{2}{c|}{Completion3D}                                                                                                                         \\ \cline{3-6} 
\multicolumn{2}{|c|}{}                                                                                                                                                                 & \multicolumn{1}{c|}{\begin{tabular}[c]{@{}c@{}}CD \\ (Value/Resolution)\end{tabular}} & \begin{tabular}[c]{@{}c@{}}EMD \\ (Value/Resolution)\end{tabular} & \multicolumn{1}{c|}{\begin{tabular}[c]{@{}c@{}}CD \\ (Value/Resolution)\end{tabular}} & \begin{tabular}[c]{@{}c@{}}EMD \\ (Value/Resolution)\end{tabular} \\ \hline
\multicolumn{1}{|c|}{\multirow{1}{*}{\begin{tabular}[c]{@{}c@{}}Point-based\\ Methods\end{tabular}}} & ASFM-Net\cite{r31}                                                                    & \multicolumn{1}{c|}{-}                                                                & -                                                                 & \multicolumn{1}{c|}{0.668/2048}                                                       & -                                                                 \\ \cline{2-6} 
\multicolumn{1}{|c|}{}                                                                                   & \begin{tabular}[c]{@{}c@{}}An End-to-End \\ Shape-Preserving\cite{r33}\end{tabular}   & \multicolumn{1}{c|}{-}                                                                & -                                                                 & \multicolumn{1}{c|}{26.900/1024}                                                        & 3.370/1024                                                         \\  \cline{2-6} 
\multicolumn{1}{|c|}{}                                                                                   & N-DPC\cite{r26}   & \multicolumn{1}{c|}{-}                                                                & -                                                                 & \multicolumn{1}{c|}{1.695/2048}                                                        & -                                                         \\ \hline
\multicolumn{1}{|c|}{\begin{tabular}[c]{@{}c@{}}Convolution-based \\ Methods\end{tabular}}                & GRNet\cite{r41}                                                                       & \multicolumn{1}{c|}{1.064/2048}                                                                & -                                                                 & \multicolumn{1}{c|}{1.046/2048}                                                       & -                                                                 \\ \cline{2-6} 
\multicolumn{1}{|c|}{}                                                                                   & SoftPoolNet\cite{r100}   & \multicolumn{1}{c|}{-}                                                                & -                                                                 & \multicolumn{1}{c|}{1.19/2048}                                                        & -                                                         \\ \hline
\multicolumn{1}{|c|}{\multirow{1}{*}{\begin{tabular}[c]{@{}c@{}}Graph-based \\ Methods\end{tabular}}}    & PRSCN\cite{r51}                                                                       & \multicolumn{1}{c|}{4.649/2048}                                                       & -                                                                 & \multicolumn{1}{c|}{-}                                                                & -                                                                 \\ \cline{2-6} 
\multicolumn{1}{|c|}{}                                                                                   & \begin{tabular}[c]{@{}c@{}}Cross-Regional \\ Attention Network\cite{r53}\end{tabular} & \multicolumn{1}{c|}{\begin{tabular}[c]{@{}c@{}}27.00/2048 \\ (6 categories \\without normalization)\end{tabular}}                                                         & 11.700/16384                                                        & \multicolumn{1}{c|}{-}                                                                & -                                                                 \\ \hline
\multicolumn{1}{|c|}{\multirow{1}{*}{\begin{tabular}[c]{@{}c@{}}GAN-based\\ Methods\end{tabular}}}       & CRN\cite{r71}                                                                         & \multicolumn{1}{c|}{2.293/2048}                                                       & -                                                                 & \multicolumn{1}{c|}{0.921/2048}                                                             & -                                                                 \\ \cline{2-6} 
\multicolumn{1}{|c|}{}                                                                                   & \begin{tabular}[c]{@{}c@{}}Cascaded Refinement \\ Network\cite{r72}\end{tabular}      & \multicolumn{1}{c|}{2.472/16384}                                                      & -                                                                 & \multicolumn{1}{c|}{0.914/16384}                                                      & -                                                                 \\  \hline
\multicolumn{1}{|c|}{\begin{tabular}[c]{@{}c@{}}VAE-based\\ Methods\end{tabular}}                        & HyperPocket\cite{r91}                                                                 & \multicolumn{1}{c|}{-}                                                                & -                                                                 & \multicolumn{1}{c|}{1.791/2048}                                                       & -                                                                 \\\hline
\multicolumn{1}{|c|}{\begin{tabular}[c]{@{}c@{}}Transformer-based\\ Methods\end{tabular}}                & SnowflakeNet\cite{r90}                                                                & \multicolumn{1}{c|}{-}                                                                & -                                                                 & \multicolumn{1}{c|}{0.760/2048}                                                      & -                                                                 \\ \cline{2-6} 
\multicolumn{1}{|c|}{}                                                                                   & PointAttN\cite{r130}   & \multicolumn{1}{c|}{-}                                                                & -                                                                 & \multicolumn{1}{c|}{0.663/2048}                                                        & -                                                         \\ \hline
\multicolumn{1}{|c|}{Other Methods}                                                                      & PMP-Net\cite{r101}                                                                     & \multicolumn{1}{c|}{-}                                                                & -                                                                 & \multicolumn{1}{c|}{0.923/2048}                                                       & -                                                                 \\ \hline
\end{tabular}%
}
\vspace{-0.5cm}
\end{table*}

\subsection{Variational autoencoders (VAEs)-based methods}

\begin{figure}[ht]
    \centering
    \includegraphics[width=12pc]{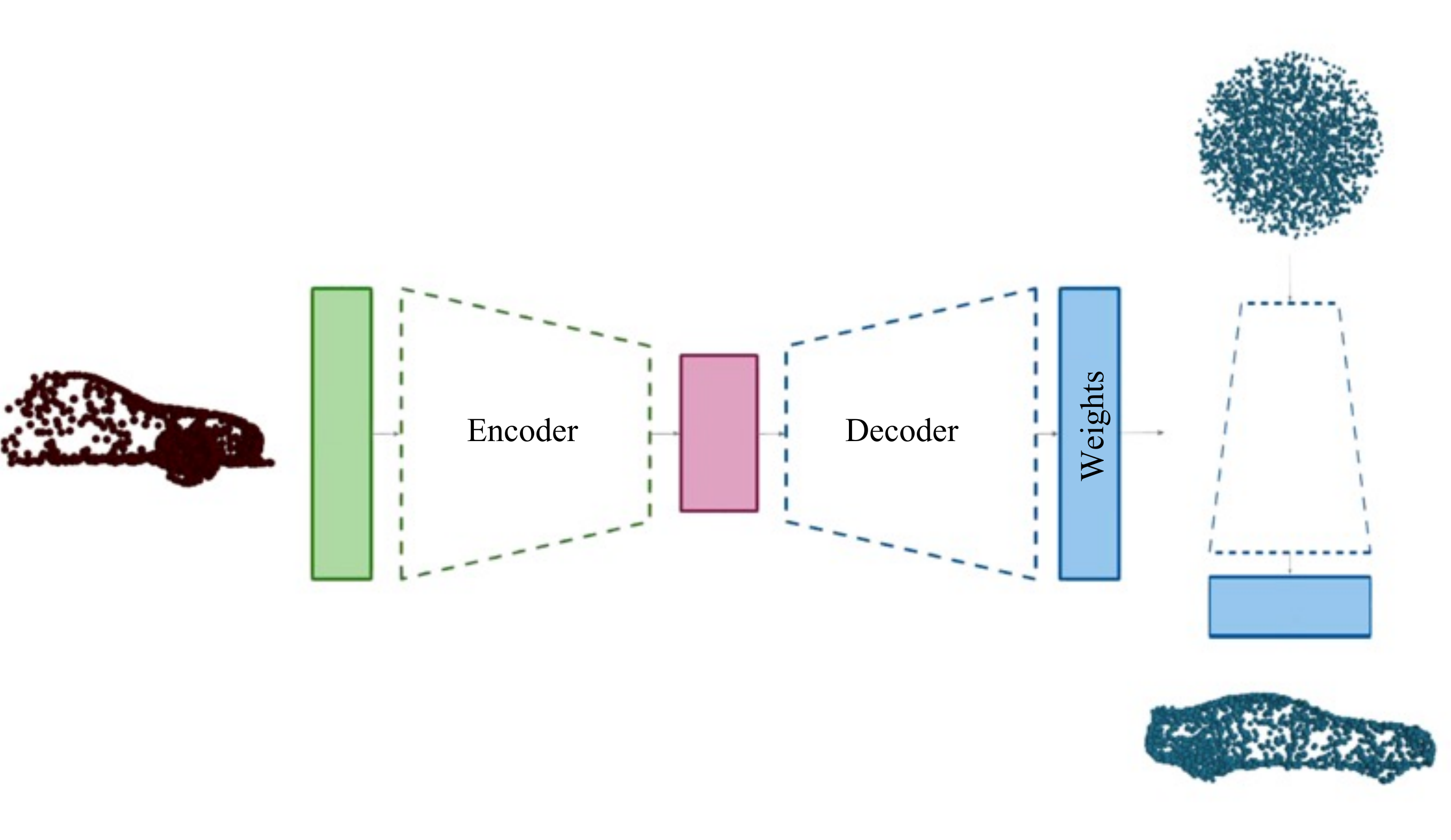}
    \caption{HyperPocket architecture with single VAE encoder \cite{r91}.}
    \vspace{-0.3cm}
\end{figure}

Classic AEs and VAEs are trained on a complete 3D object. The model's weight is then determined to generate a latent representation of incomplete data. And the generative model completes partial inputs in conditional generative network settings. The completion production is based on the learning mode distribution extracted from the complete shape.

Spurek et al. \cite{r92} introduced a Variational Autoencoder architecture named HyperPocket, which is capable of disentangling latent representations and thus generating multiple variants of completed 3D point clouds (Fig. 16). The point cloud processing was split into two unconnected data streams and utilized a hyper network paradigm to fill what are known as pockets of space left by missing object parts. Pan et al. \cite{r91} devised a Variational Relational Point Completion Network (VRCNet) to leverage a dual-path unit and a VAE-based relational enhancement module for probabilistic modeling. And they also designed multiple relational modules that could efficiently utilize and integrate multi-scale point information, including the Point Self-Attention Kernel and the Point Selective Kernel Unit. Zamorski et al. \cite{r93} presented an application of three generative modeling approaches and tested the architectures of AE, VAE, and Adversarial Autoencoder both quantitatively and qualitatively. Besides, they introduced a method that uses the extended PointNet model (Double PointNet) to manipulate points based on both local features and the global shape. AutoSDF \cite{r132} proposed an autoregressive prior for 3D shapes to solve multi-modal 3D tasks such as shape completion, reconstruction, and generation. The distribution over 3D shapes is modeled as a nonsequential autoregressive distribution over a discretized, low-dimensional, symbolic grid-like latent representation of 3D shapes. This enables the network to represent distributions over 3D shapes conditioned on information from an arbitrary set of spatially anchored query locations and thus perform shape completion in such arbitrary settings.

However, the advantages and disadvantages of VAE-based methods can be concluded as follow:
\begin{itemize}
    \item The training of VAE-based methods is more stable, compared with other (3D) generative models.
    \item The latent representations of VAE can be manipulated to control the shapes of the generated point cloud \cite{r92}.
    \item The quality of VAE-based methods cannot be comparable to the GAN-based generative models, while the diversity of VAE-based methods is superior than GAN-based methods. Because of the injected noise and imperfect element-wise measures such as the squared error, the generated 3D shapes might not be smooth enough.
    \item The VAE-based methods are enhanced by AutoSDF\cite{r132}, which is a VQ-VAE-like model. The great success of VQ-VAE\cite{r131} in image generation will also boost the VAE-based methods on point cloud completion.
\end{itemize}

\subsection{Transformer-based methods}

Transformer \cite{r83} was firstly proposed for encoding sentences in natural language processing, and after that became popular in the areas of 2D computer vision (CV) \cite{r84, r85}. Pioneered by PCT \cite{r86}, Pointformer \cite{r87}, and PointTransformer \cite{r88}, the transformer has began its journey in point cloud process. 

\begin{figure}[ht]
    \centering
    \includegraphics[width=21pc]{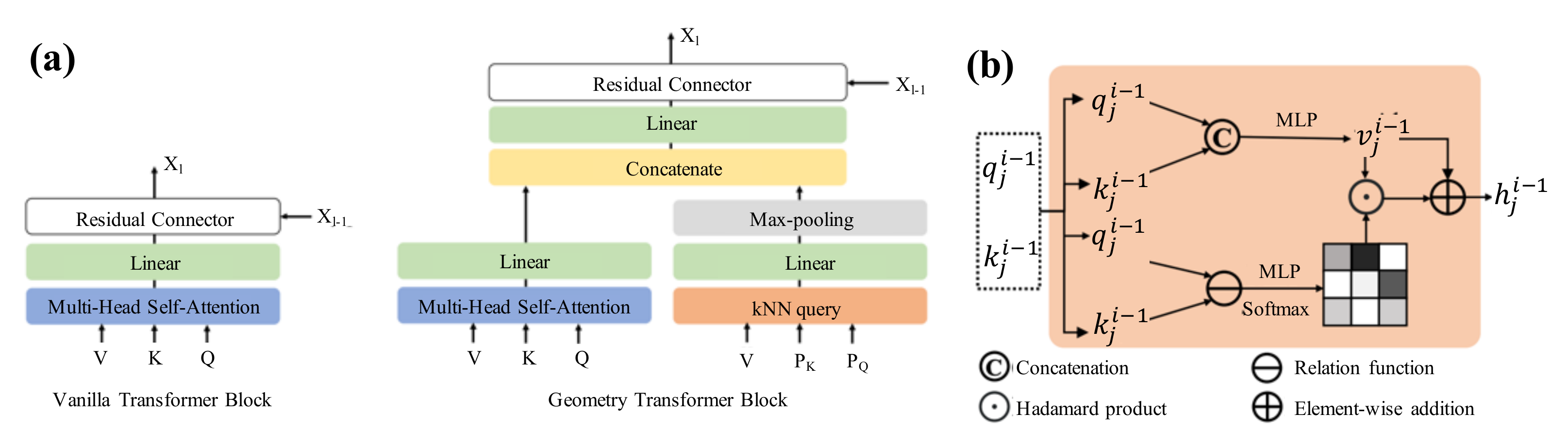}
    \caption{(a) Compares of the transformer block and the geometry-aware transformer block \cite{r89}; (b) The detailed structure of skip-transformer \cite{r90}.}
    \vspace{-0.3cm}
\end{figure}

By the merits of the representation learning ability of transformer, Yu et al. \cite{r89} regarded point cloud completion as a set-to-set translation issue and proposed a transformer encoder-decoder structure for point cloud completion. By representing the point cloud as a set of disordered points with position embeddings, The point cloud can be converted to a series of point proxies. The transformer was employed for the point cloud generation. To facilitate transformers to use better the inductive bias of 3D geometric structures of the point cloud, they further design a geometry-aware block that explicitly simulates local geometric relations (Fig. 17a).

Rather than utilizing the representation learning ability of transformer, Xiang et al. \cite{r90} devised SnowflakeNet with Snowflake Point Deconvolution (SPD), applying transformer-based structure to the decoding procedure. The SnowflakeNet models the generation of complete point clouds as the snowflake-like growth of points in 3D space. After each SPD, child points are gradually generated by splitting their parent points. The idea of revealing geometry details is to introduce a skip-transformer in SPD to learn point splitting modes that can best suit local regions. Skip-transformer utilizes an attention mechanism to summarize the splitting patterns used in the previous SPD layer, resulting in the splitting of the current SPD layer. The locally compact, structured point cloud produced by SPD can accurately capture the structure features of 3D shape in local patches, enabling the network to predict highly detailed geometries (Fig. 17b).

Moreover, Lin et al. \cite{r122} presented PCTMA-Net, where the Transformer’s attention mechanism can extract the local context within a point cloud and exploit its incomplete local structure details. A morphing-atlas-based point generation network fully utilizes the extracted point Transformer feature to predict the missing region using charts defined on the shape. Improved from PMP-Net \cite{r101}, PMP-Net++ \cite{r13} introduced a transformer-enhanced representation learning network, significantly improving completion performance.

However, there are some limitations of transformer-based models.
\begin{itemize}
    \item Due to the amount of the transformer parameter, the model is too large to deploy on devices compared with other methods.
    \item Except for the visual interpretation of attention in SA-Net \cite{r58}, the mechanism of transformer for enhanced performance is hard to interpret.
\end{itemize}

% Please add the following required packages to your document preamble:
% \usepackage{multirow}
% \usepackage{graphicx}
\begin{table*}[htbp]\footnotesize
\caption{Comparison of 3D point clouds completion performances on the KITTI. 'MMD' represents the Minimal Matching Distance and 'FD' represents the Fidelity Distance. The '-' stands for the performances are unreachable.}
\centering
\renewcommand{\arraystretch}{1.1}
{%
\begin{tabular}{|cc|ccc|}
\hline
\multicolumn{2}{|c|}{\multirow{1}{*}{Methods}}                                                                            & \multicolumn{3}{c|}{KITTI}                                                                                                                                                                          \\ \cline{3-5} 
\multicolumn{2}{|c|}{}                                                                                                    & \multicolumn{1}{c|}{\begin{tabular}[c]{@{}c@{}}MMD\\  (Value/Resolution)\end{tabular}} & \multicolumn{1}{c|}{\begin{tabular}[c]{@{}c@{}}FD\\ (Value/Resolution)\end{tabular}} & Consistency(Value/Resolution) \\ \hline
\multicolumn{1}{|c|}{\multirow{1}{*}{\begin{tabular}[c]{@{}c@{}}Point-based \\ Methods\end{tabular}}}       & PCN\cite{r17}         & \multicolumn{1}{c|}{0.0185/16384}                                                      & \multicolumn{1}{c|}{0.028/16384}                                                     & 0.01163/16384                 \\ \cline{2-5} 
\multicolumn{1}{|c|}{}                                                                                      & ASFM-Net\cite{r31}    & \multicolumn{1}{c|}{-}                                                                 & \multicolumn{1}{c|}{0.014/4096}                                                      & 0.020/4096                    \\ \cline{2-5} 
\multicolumn{1}{|c|}{} &
ASHF-Net\cite{r59}    & \multicolumn{1}{c|}{0.541/16384}                                                       & \multicolumn{1}{c|}{0.773/16384}                                                     & 0.298/16384                   \\ \hline
\multicolumn{1}{|c|}{\multirow{1}{*}{\begin{tabular}[c]{@{}c@{}}Convolution-based \\ Methods\end{tabular}}} & GRNet\cite{r41}       & \multicolumn{1}{c|}{-}                                                                 & \multicolumn{1}{c|}{-}                                                               & 0.313/16384                   \\ \cline{2-5} 
\multicolumn{1}{|c|}{}                                                                                      & SoftPoolNet\cite{r100} & \multicolumn{1}{c|}{0.01465/16384}                                                     & \multicolumn{1}{c|}{0.02171/16384}                                                   & 0.00992/16384                 \\ \hline

\multicolumn{1}{|c|}{\multirow{1}{*}{GAN-based Methods}}                                                    & MMD-GAN\cite{r70}     & \multicolumn{1}{c|}{-}                                                                 & \multicolumn{1}{c|}{0.034/2048}                                                      & -                             \\ \cline{2-5} 
\multicolumn{1}{|c|}{}                                                                                      & NSFA\cite{r35}    & \multicolumn{1}{c|}{-}                                                                 & \multicolumn{1}{c|}{0.0261/16384}                                                    & -                             \\ \cline{2-5} 
\multicolumn{1}{|c|}{}                                                                                      & SpareNet    & \multicolumn{1}{c|}{0.368/2048}                                                        & \multicolumn{1}{c|}{1.461/2048}                                                      & 0.249/2048                    \\ \hline
\multicolumn{1}{|c|}{\begin{tabular}[c]{@{}c@{}}Transformer-based \\ Methods\end{tabular}}                  & PoinTr\cite{r89}      & \multicolumn{1}{c|}{-}                                                                 & \multicolumn{1}{c|}{0.526/8192}                                                      & -                             \\ \hline
\end{tabular}%
}
\vspace{-0.5cm}
\end{table*}

\subsection{Other methods}
In addition to the methods mentioned above, researchers have also carried out studies on up-sampling and pre-train methods.

Wen et al. \cite{r101} designed PMP-Net to complete the point cloud by moving every point in the incomplete input to ensure the shortest total distance of the point moving path (PMP). Therefore, PMP-Net predicts the unique PMP of each point based on the constraint of the total point movement distance. Kim et al. \cite{r102} introduced a shape completion framework to preserve both the global contexts and the local features, in which a Symmetry-Aware Up-sampling Module (SAUM) was devised to preserve the geometric details and take advantage of the symmetry of shape completion. Kusner et al. \cite{r103} developed a pre-training mechanism named Occlusion Completion (OcCo), which works by shielding the occluded points from observations from different camera views and then optimizing the completion model. In this way, This method learns a pre-trained representation that could recognize the inherent visual constraints embedded in a real point cloud.

\section{Comparison}
This section summarizes the results of the latest methods on several datasets. We will compare the performance of these approaches and provide some advice for future works. The results come from the original papers. Thus the resolutions and datasets setting are varied. If the type of CD or datasets settings are not specified, the performance can be compared under the same resolution.
% Please add the following required packages to your document preamble:
% \usepackage{multirow}
% \usepackage{graphicx}
\begin{table*}[htbp]\scriptsize
\caption{Summarization of milestone DL networks based on the point cloud processing methods.}
\centering
\setlength\tabcolsep{1pt}
\renewcommand{\arraystretch}{1.03}
{%
\begin{tabular}{|cc|c|c|}
\hline
\multicolumn{2}{|c|}{Methods}                                                                                                                                                           & Highlights                                                                                                                                                                                                                                                                             & limitation                                                                                                                                                                                                         \\ \hline
\multicolumn{1}{|c|}{\multirow{1}{*}{\begin{tabular}[c]{@{}c@{}}Point-based\\  methods\end{tabular}}}       & AltasNet\cite{r12}                                                                  & \begin{tabular}[c]{@{}c@{}}AtlasNet regards a 3D shape as a collection of parametric \\ surface elements and infers a surface representation.\end{tabular}                                                                                                                             & \begin{tabular}[c]{@{}c@{}}The reconstruction largely depends \\ on the reiterated many times.\end{tabular}                                                                                                           \\ \cline{2-4} 
\multicolumn{1}{|c|}{}                                                                                      & MSN\cite{r22}                                                                       & \begin{tabular}[c]{@{}c@{}}MSN predicts a set of parametric surface elements and \\ undergoes a combination with the partial input by a sampling algorithm.\end{tabular}                                                                                                               & \begin{tabular}[c]{@{}c@{}}Fail to generate fine-grained \\ details of object shape.\end{tabular}                                                                                                                     \\ \cline{2-4} 
\multicolumn{1}{|c|}{}                                                                                      & PCN\cite{r17}                                                                      & \begin{tabular}[c]{@{}c@{}}Combing the fully connected network and FoldingNet, \\ PCN performs the coarse-to-fine completion.\end{tabular}                                                                                                                                             & Incapable of synthesizing shape details.                                                                                                                                                                              \\ \cline{2-4} 
\multicolumn{1}{|c|}{}                                                                                      & ASFM-Net\cite{r31}                                                                  & \begin{tabular}[c]{@{}c@{}}ASFM-Net is an asymmetrical Siamese auto-encoder \\ model to learn a shape prior information.\end{tabular}                                                                                                                                                  & \begin{tabular}[c]{@{}c@{}}The visualization performance of \\ completion can be further improved.\end{tabular}                                                                                                       \\ \cline{2-4} 
\multicolumn{1}{|c|}{}                                                                                      & SK-PCN\cite{r24}                                                                  & \begin{tabular}[c]{@{}c@{}}SK-PCN predicts the 3D skeleton to acquire the global structure \\ and completes the surface by learning skeletal points' displacements.\end{tabular}                                                                                                       & \begin{tabular}[c]{@{}c@{}}The meso-skeleton only focus\\  on the overall shapes.\end{tabular}                                                                                                                        \\ \cline{2-4} 
\multicolumn{1}{|c|}{}  & FoldingNet\cite{r55}                                                                & \begin{tabular}[c]{@{}c@{}}A two-stage generation process is commonly used to \\ assume that 3D objects can be recovered from a 2D-manifold.\end{tabular}                                                                                                                              & \begin{tabular}[c]{@{}c@{}}The implicit intermediate is hard to \\ be constrained explicitly.\end{tabular}                                                                                                            \\ \cline{2-4} 
\multicolumn{1}{|c|}{}                                                                                      & SA-Net\cite{r58}                                                                    & \begin{tabular}[c]{@{}c@{}}SA-Net proposes hierarchical folding in the multi-stage \\ points generation decoder.\end{tabular}                                                                                                                                                          & \begin{tabular}[c]{@{}c@{}}It is difficult to interpret and constrain the \\ implicit representation of the target \\ shape from the intermediate layer \\ to help refine the shape in the local region.\end{tabular} \\ \cline{2-4} 
\multicolumn{1}{|c|}{}                                                                                      & ASHF-Net\cite{r59}                                                                  & \begin{tabular}[c]{@{}c@{}}ASHF-Net proposes a hierarchical folding decoder with the gated\\  skip-attention and multi-resolution completion target to exploit \\ the local structure details of the incomplete inputs.\end{tabular}                                                   & \begin{tabular}[c]{@{}c@{}}The surface of the results is not smooth. Unstructured \\predictions are obtained by the decoder\cite{r90}.  \end{tabular}                                                                                                                                                                      \\ \hline
\multicolumn{1}{|c|}{\multirow{1}{*}{\begin{tabular}[c]{@{}c@{}}Convolution-based\\  methods\end{tabular}}} & GRNet\cite{r41}                                                                     & \begin{tabular}[c]{@{}c@{}}GRNet introduces 3D grids as intermediate representations \\ to regularize unordered point clouds.\end{tabular}                                                                                                                                             & \begin{tabular}[c]{@{}c@{}}It is still subject to the resolution. Higher resolution\\ will bring about great computational cost \end{tabular}                                                                                                                                                                                \\ \cline{2-4} 
\multicolumn{1}{|c|}{}                                                                                      & VE-PCN\cite{r119}                                                                    & \begin{tabular}[c]{@{}c@{}}VE-PCN incorporates the structure information into \\ the shape completion by leveraging edge generation.\end{tabular}                                                                                                                                     & \begin{tabular}[c]{@{}c@{}}The edges only focus on high \\ frequency components.\end{tabular}                                                                                                                              \\ \hline
\multicolumn{1}{|c|}{\multirow{1}{*}{\begin{tabular}[c]{@{}c@{}}Graph-based\\ \\  methods\end{tabular}}}    & PRSCN\cite{r51}                                                                     & \begin{tabular}[c]{@{}c@{}}PRSCN proposes a Point Rank Sampling to select feature points \\ and eliminate the influence of outlier points. Point Rank Sampling \\ pays more attention to the local relative importance and generates \\ high fidelity geometrical shapes.\end{tabular} & \begin{tabular}[c]{@{}c@{}}It can only predict the missing parts.\\ The Point Rank Sampling (PRS) module is \\ susceptible to the density distribution. The \\ generalization ability needs to be improved, facing \\ data from other domains with different data \\ distribution and object patterns \end{tabular}                                                                                                                                                                               \\ \cline{2-4} 
\multicolumn{1}{|c|}{}                                                                                      & ECG\cite{r48}                                                                       & \begin{tabular}[c]{@{}c@{}}ECG applies several modules with graph convolutions \\ for edge-aware feature learning and/or preserving.\end{tabular}                                                                                                                                      & \begin{tabular}[c]{@{}c@{}}There are noises around \\ the completion result.\end{tabular}                                                                                                                             \\ \cline{2-4} 
\multicolumn{1}{|c|}{}                                                                                      & DCG\cite{r116}                                                                       & \begin{tabular}[c]{@{}c@{}}DCG employs the point set auto-encoder first to produce a sparsely coarse \\ shape and then refines it by encoding neighborhood connectivity\\  on a graph representation.\end{tabular}                                                                     & \begin{tabular}[c]{@{}c@{}}The performance can be further improved,\\ due to the noises around the final output.\\And the uneveenly distributed points might lead\\ to high EMD.\end{tabular}                                                                                                                                   \\ \cline{2-4} 
\multicolumn{1}{|c|}{}                                                                                      & CRA-Net\cite{r52}                                                                   & \begin{tabular}[c]{@{}c@{}}The CRA module facilitates each local feature vector to search the regions\\  within a fully-connected graph and selectively absorbs other local features\\  based on their relationships with graph convolution.\end{tabular}                              & \begin{tabular}[c]{@{}c@{}}Insensitive to edge information. Failed to fully\\ exploit the local features, which is soloved by \cite{r53}\end{tabular}                                                                                                                                                                                       \\ \hline
\multicolumn{1}{|c|}{\multirow{1}{*}{\begin{tabular}[c]{@{}c@{}}View-based\\  methods\end{tabular}}}                                                                                        & ViPC\cite{r44}                                                                      & \begin{tabular}[c]{@{}c@{}}An extra single-view image explicitly provides the \\ global structural prior information for completion.\end{tabular}                                                                                                                                      & \begin{tabular}[c]{@{}c@{}}The surface of the results is not that smooth.\\ And the performance is largely depended on the\\ angles of images.\end{tabular}    \\  \hline 

\multicolumn{1}{|c|}{\multirow{1}{*}{\begin{tabular}[c]{@{}c@{}}GAN-based\\  methods\end{tabular}}}      & PF-Net\cite{r82}                                                                    & \begin{tabular}[c]{@{}c@{}}PF-Net uses a feature-points-based multi-scale generating network \\ and combines multi-stage completion loss and adversarial loss \\ to generate realistic missing regions.\end{tabular}                                                                   & \begin{tabular}[c]{@{}c@{}}PF-Net neglects effective feature fusion,\\ resulting the shape lacks  local geometric details.\end{tabular}                                                                                                                         \\ \cline{2-4} 
\multicolumn{1}{|c|}{}                                                                                      & ShapeInversion\cite{r75}                                                            & \begin{tabular}[c]{@{}c@{}}ShapeInversion addresses the domain gaps between virtual\\  and real-world partial scans and various simulated partial shapes\\ through GAN inversion.\end{tabular}                                                                                         & \begin{tabular}[c]{@{}c@{}}Both shape completion and manipulation\\  are conducted on a model pre-trained with\\  a single category\end{tabular}                                                                      \\ \cline{2-4} 
\multicolumn{1}{|c|}{}                                                                                      & \begin{tabular}[c]{@{}c@{}}MMD-GAN\\ (Learning Shape Priors)\end{tabular} & \begin{tabular}[c]{@{}c@{}}MMD-GAN proposes 3D feature alignment methods to \\ learn the shape priors from complete and partial point clouds.\end{tabular}                                                                                                                             & \begin{tabular}[c]{@{}c@{}}Insensitive to edge information.\\  The performance might be further enhanced by training\\ a model in one class as \cite{r75}.\end{tabular}                                                                                                                                                                                       \\ \cline{2-4} 
\multicolumn{1}{|c|}{}                                                                                      & NSFA\cite{r35}                                                                      & \begin{tabular}[c]{@{}c@{}}NSFA proposes two separated feature aggregation namely GLFA \\ and RFA, considers the existing known part and the missing part separately.\end{tabular}                                                                                                     & \begin{tabular}[c]{@{}c@{}}Suffer from the information loss of\\  structure details.\end{tabular}                                                                                                                     \\ \cline{2-4} 
\multicolumn{1}{|c|}{}                                                                                      & \begin{tabular}[c]{@{}c@{}}Cascaded Refinement\\ Network\cite{r71}\end{tabular}     & \begin{tabular}[c]{@{}c@{}}The generator of CRN  is a cascaded refinement network, \\ exploiting the details of the partial inputs and synthesizing \\ the missing parts with high quality.\end{tabular}                                                                               &  \begin{tabular}[c]{@{}c@{}}The performance can be further enhanced. \\ There are a few pairs of partial and complete point \\ cloud with unmatched scales in CRN dataset.\end{tabular}                                                                                                                                                                             \\ \hline
\multicolumn{1}{|c|}{\multirow{1}{*}{\begin{tabular}[c]{@{}c@{}}Transformer-based\\  methods\end{tabular}}} & PoinTr\cite{r89}                                                                    & \begin{tabular}[c]{@{}c@{}}PoinTr regards point cloud completion as a set-to-set \\ translation issue and employs a transformer encoder-decoder architecture.\end{tabular}                                                                                                             & \begin{tabular}[c]{@{}c@{}}The model is relatively large due to \\ transformer's large number of parameters.\end{tabular}                                                                                             \\ \cline{2-4} 
\multicolumn{1}{|c|}{}                                                                                      & SnowflakeNet\cite{r90}                                                              & \begin{tabular}[c]{@{}c@{}}SnowflakeNet introduces a skip-transformer to learn splitting\\  patterns  in Snowflake Point Deconvolution for progressively \\ increasing the number of points.\end{tabular}                                                                              & \begin{tabular}[c]{@{}c@{}}The resulting point cloud is not evenly distributed. \\ It focues more on structured output rather than\\ topological information\cite{r139}.  \end{tabular}                                                                                                                        \\ \hline
\end{tabular}%
}
\vspace{-0.5cm}
\end{table*}

% Please add the following required packages to your document preamble:
% \usepackage{graphicx}
\begin{table}[htbp]\footnotesize
\centering
\setlength\tabcolsep{0.8pt}
\caption{\textbf{Complexity analysis.} The number of the parameter (Params) and theoretical computation cost (FLOPs) of existing methods are reported. The average L2 Chamfer distances of all categories in ShapeNet-55 and unseen categories in ShapeNet34 are also provided.}
{%
\begin{tabular}{c|cc|cc}
\toprule[1.5pt]
Methods      & Param.  & FLOPS   & \begin{tabular}[c]{@{}c@{}}ShapeNet55 \\ -CD-$\ell_2$\end{tabular}  & \begin{tabular}[c]{@{}c@{}}ShapeNet34 \\ -CD-$\ell_2$\end{tabular} \\ \midrule[1pt]
FoldingNet \cite{r55}   & 2.30 M  & 27.58 G & 3.12             & 3.62             \\
PCN \cite{r18}         & 5.04 M  & 15.25 G & 2.66             & 3.85             \\
TopNet \cite{r16}      & 5.76 M  & 6.72 G  & 2.91             & 3.50             \\
PF-Net \cite{r82}       & 73.05 M & 4.96 G  & 5.22             & 8.16             \\
GRNet \cite{r41}       & 73.15 M & 40.44 G & 1.97             & 2.99             \\
ECG \cite{r48}         & 13.77 M & 9.32 G  & 1.76             & 2.09             \\
CRN  \cite{r71}        & 5.00 M  & 8.41 G  & 2.01             & 3.30             \\
ASFM-Net \cite{r31}    & 10.75 M & 17.41 G & 1.70             & 2.67             \\
PMPNet\cite{r101}       & 5.18 M  & 3.05 G  & 1.47             & 2.15             \\
SnowflakeNet\cite{r90} & 18.43 M & 5.20 G  & 1.21             & 2.20             \\
PoinTr \cite{r89}       & 30.9 M  & 10.41 G & 1.07             & 2.05             \\ \bottomrule[1.5pt]
\end{tabular}%
}
\vspace{-0.5cm}
\end{table}

\subsection{Summary of the performance on PCN, ModelNet, and Completion3D with ground truth provided.}

%\begin{figure*}[htbp]
%    \centering
%    \includegraphics[width=43pc]{figureb6.png}
%    \caption{Qualitative completion %results on ShapeNet-55 \cite{r89}, which is drived from ShapeNet.}
%\end{figure*}
The PCN is the most commonly utilized dataset for 3D shape completion. These three datasets all belong to synthetic benchmarks. As shown in Table II, III, V, and Fig. s2 (see the Supplementary Material), there are the results performed by various methods, and some inferences could be drawn as follow:

\begin{itemize}
	\item The point-based models with MLP, integrated as the basic unit, are widely utilized to learn the point-wise information.
    \item The graph-based and GAN-based networks can fulfill excellent results on completing 3D point clouds. More attention needs to be paid to the combination of these two methods.
    \item Transformer-based models have recently attracted more attention because of their powerful ability to process irregular data. The SOTA methods could be credited to the latest SnowflakeNet. Nevertheless, extending transformer-based models into the spectral domain remains a challenge.
\end{itemize}

\subsection{Summary of the performance on KITTI without ground truth provided.}

%\begin{figure*}[]
%    \centering
%    \includegraphics[width=43pc]{figureb7.png}
%    \caption{Qualitative completion results on KITTI \cite{r41}.}
%    \label{fig:my_label}
%\end{figure*}

Due to the missing ground truth, the performance in the KITTI dataset is generally obtained by testing the model that trained directly on the training set of "cars" in ShapeNet. And several methods fine-tune their model on ShapeNetCars as \cite{r41} while testing the model on KITTI. Tables IV, V, and Fig. s2 (see the Supplementary Material) give the performance achieved by numerous methods on the KITTI, and from which some observations could be listed as follow:

\begin{itemize}
	\item The KITTI dataset is derived from the real-world scans. The intrinsics challenges such as no ground truth provided and ultimately sparse in some instances bring difficulties to point cloud completion.
	\item As shown in Table III, the Point-based, GAN-based, and Transformer-based methods all have achieved completion effects facing such challenges.
	\item Besides, some works \cite{r68, r75} are devised for point cloud in the real world, and more effective should be paid on these directions. 
\end{itemize}

\subsection{Complexity analysis and generalization performance}
To further gain insight into model performances,  further analysis of the parameter (Params) and theoretical computation cost (FLOPs) are conducted to compare the complexity of the model and the time-consuming situation. As shown in the second and third columns of Table VI, it can be found that FoldingNet \cite{r55} owns the smallest number of parameters, while the GRNet \cite{r41} and PF-Net \cite{r82} possess a relatively great number of parameters due to the complex architecture. On the other hand, PMPNet \cite{r101} has the lowest computational cost while GRNet still has the highest one since the operator on their grids needs more parameters and computation. It is noticed that the number of parameters in SnowflakeNet \cite{r90} and PoinTr \cite{r89} is also relatively high because of the attention mechanism.

Furthermore, generalization results of models are also compared in Table VI since the performance in unknown categories is also another critical metric. Because ShapeNet55 and ShapeNet34 are proposed to measure the generalization performance, the models are trained in the seen 34 categories and evaluated on the unseen 21 categories. As we can see in the last two columns of Table VI, PoinTr \cite{r89} performs well in ShapeNet55 and ShapeNet34, proving the most remarkable generalization of PoinTr.

\section{Applications}
The point cloud completion is a vital technology in many applications and has accumulated several achievements. Therefore, in this section, the applications of point cloud completion in numerous fields will be introduced. We will discuss the applications where point cloud completion can be utilized at current firstly. Then, four major applications where point cloud completion will be applied are also reviewed.

\subsection{Current applications}

This paper discusses the techniques of point cloud completion are object-level completion. Therefore, these methods are excellent in single object completion in real-world scenes, for example, reconstructions of a chair in real-world scenes produced by the HyperPocket \cite{r91} model. Relationship-based methods \cite{r115} complete partial scans of pairwise scenes by conditioning the completion of each object with the partial scan of the other object. Both of them can complete the real-world scenes for the convenience of indoor robots. Further, Gu et al. \cite{r68} proposed a weakly supervised method to estimate both 3D canonical shape and 6-DoF pose for alignment, given multiple partial observations associated with the same instance. This method enables the completion toward objects in the wild, which might benefit the high-precision of 3D object detection or tracking in outdoor scenes. Considering many works being powerful for synthesis datasets, significant efforts should be made to enable object-level point cloud completion in more practical scenarios.

%\begin{figure}[h]
%    \centering
%    \includegraphics[width=21pc]{figure12.png}
%    \caption{An illustration of point clouds in the construction.}
%    \label{fig:my_label}
%\end{figure}

\subsection{Construction}
Because of the enormous benefits, the completed point cloud is urgently needed by industries to enhance productivity, such as the manufacturing \cite{r104, r105} and construction industry \cite{r106, r107}. For instance, as shown in Fig. 18a \cite{r142}, the point cloud was acquired at a precast concrete manufacturing plant. Compared with traditional measurement approaches, such as manual or other measurements based on equipment, the point cloud data captured by sensors has the merits of a higher measurement rate and higher measurement accuracy.

%\begin{figure}[h]
%    \centering
%    \includegraphics[width=21pc]{figure13.png}
%    \caption{Point cloud of fully mechanized mining equipment.}
%    \label{fig:my_label}
%\end{figure}

\begin{figure}[ht]
    \centering
    \includegraphics[width=18pc]{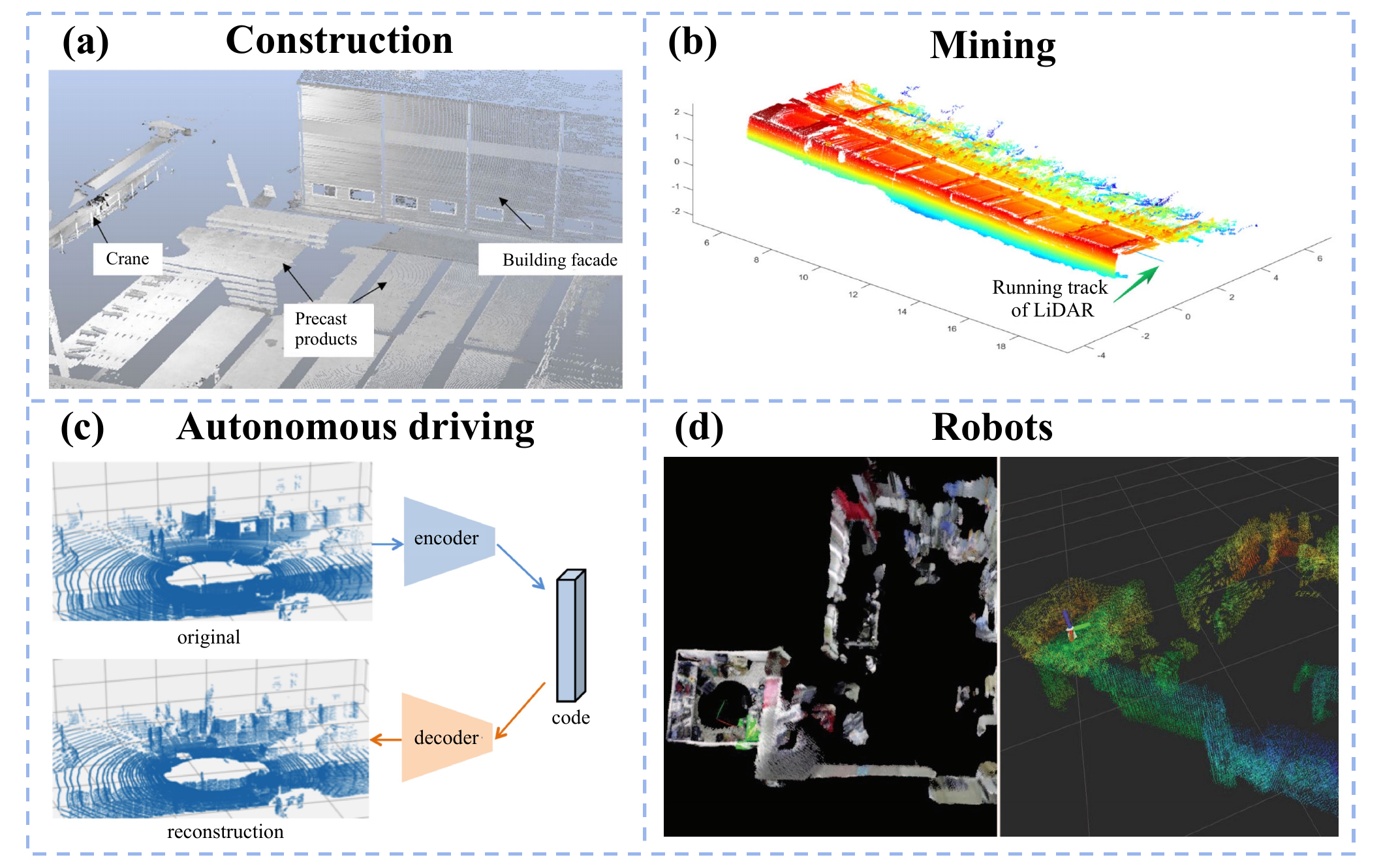}
    \caption{Point cloud completion in (a) construction \cite{r142}, (b) fully mechanized mining equipment \cite{r143}, (c) 3D reconstruction \cite{r110}, and (d) localization of autonomous indoor robots \cite{r112}.}
    \vspace{-0.5cm}
\end{figure}

\subsection{Mining space}
Nowadays, 3D point cloud processing technology has been applied frequently in mining. For example, the national robotics engineering center of the United States can successfully draw a high-precision, 3D map of underground roads using the point cloud data obtained by a 3D laser scanner and then propose an intelligent mining model based on the 3D map \cite{r113}. The 3D point cloud data is used to describe and draw the whole fully-mechanized mining face, accurately and intuitively reflecting the spatial position relationship between the coal wall and fully-mechanized mining equipment. This method provides directional information for the scraper conveyor to adjust the displacement of the hydraulic support in time (Fig. 18b) \cite{r143}. Significantly, the complete point cloud will provide more accurate information for mining space. 

%\begin{figure}[h]
%    \centering
%    \includegraphics[width=21pc]{figure14.png}
%    \caption{The target of 3D reconstruction is to find a compact 3D representation that retains the capacity for reconstruction.}
%    \label{fig:my_label}
%\end{figure}

\subsection{Autonomous driving}
On the one hand, the main task of autopilot is to find a compact 3D point cloud representation and maintain the capacity for reconstruction; As shown in Fig. 18c, reconstruction helps to store data in autonomous driving. Because every autonomous vehicle (AV) has to store high-definition maps and collect real-time LiDAR sweeps, data storage would be expensive for a large fleet of AVs. While no mature compression standard is available to deal with large-scale open scene 3D point clouds \cite{r110}, reconstruction technology can provide 3D point cloud compression and reduce the cost of data storage. Point cloud completion can be used for reconstruction to obtain a higher quality point cloud.

On the other hand, the production of high-definition maps is relatively expensive and impractical for every scene. Hence, semantic scene completion is proposed to complete the sparse LiDAR sweeps. In this area, convolution-based methods have been widely applied. However, the details of scenes are missing due to the voxelization of the point cloud \cite{r126,r127,r128}. To tackle this issue, the completion of sparse LiDAR sweeps through point-based methods might be a solution.

Images, point clouds, and radar data could be combined to produce precise, geo-referenced, and information-rich cues for AVs' navigation and decision-making \cite{r111}. Data from low-end LiDAR and high-end LiDAR are also fused. At the same time, there are some difficulties in merging these data. Most importantly, in the fusion of cross-source data, the sparsity of point cloud leads to inconsistent and missing data. Therefore, the point cloud completion can be utilized to tackle the sparsity of real-time LiDAR sweeps.

%\begin{figure}[h]
%    \centering
%    \includegraphics[width=21pc]{figure15.png}
%    \caption{Localization of autonomous robot in indoor 3D environment. Left: The green area represents the final alignment of the transformation matrix estimated on a given global map. Right: Robot pose on ROS-RVIz's map display}
%    \label{fig:my_label}
%\end{figure}

\subsection{Robotics}
Point cloud technology has been widely used in robotics in recent years. Localization and mapping are crucial for autonomous mobile robot navigation in unknown environments. \textbf{Localization} A accurate 6-degree of freedom (6-DoF) pose is ideal for unmanned aerial vehicles (UAVs) or humanoid robots and robots performing missions. However, localization with RGB-D cameras in a 3D environment still presents some challenges: (1) Robots often take a long time to orient themselves in a 3D environment; (2) There are changes in the 3D environment. For instance, as shown in Fig. 18d, Luo et al. presented a method to determine the 6-DoF global positioning of the robot without a given initial pose, which is dubbed the Fast Scene Recognition and Alignment (FSRA) system \cite{r112}. \textbf{Mapping} More attention has been paid to the three-dimensional environment point cloud maps in public and engineering space. 3D environment maps are helpful for autonomous mobile robots in indoor environments without GPS. However, these precise locations and mappings still require a complete point cloud as a prerequisite.

\section{Future direction and open questions}
Based on the above discussions, there are two problems to be solved : (1) To achieve high precision and robust completion by overcoming the above challenges. (2) Fast operation speed and high accuracy guarantee. In this part, we propose several future research directions to enhance the performance of DL-based point cloud network as follows:
\begin{itemize}
    \item Unlike images, point cloud cannot find the corresponding "points" in the paired point cloud. If this issue can be tackled, many methods in image generation can also be applied in point cloud completion.
    \item View-based methods can be utilized to be combined with recently popular Neural Radiance Fields (NeRF) to reconstruct more realistic 3D shapes from the images.
    \item Novel generative models need to be integrated into point cloud completion, such as the diffusion model. Although a few works \cite{r141} have been made, the speed of generating samples is demanded to be improved.
    \item Moreover, establishing the novel loss functions is a significant challenge to be solved in the future, such as DCD loss \cite{r125}.
    \item Although the DL-based point cloud completion has achieved impressive results, almost all the existing networks are conducted in the current datasets, such as PCN, ModelNet, and Completion3D. These datasets are derived from the CAD. Therefore, it is urgent to develop new datasets captured in the real world to make the networks more robust in the wild. Although unsupervised methods \cite{r75, r94, r95} have been developed, more effects should be paid on it because the captured point clouds from real-world could not obtain the ground truth. 
    \item Due to the disorder and irregularity of point cloud, the early processing of point cloud is mainly voxelization. Still, this processing method will lead to the loss of effective information of the point cloud and increased computational complexity. Although feature extraction networks have been designed, such as PointNet and GCN, more effects should be paid to feature learning. In the decoder design, there are only fully connected networks, FoldingNet, and the newly-proposed transformer-based decoder network.
    \item  Although there are remarkable achievements in 3-D DL models, including PointNet \cite{r10}, PointNet++ \cite{r11}, PointCNN \cite{r38}, DGCNN \cite{r45}, FoldingNet \cite{r55}, PF-Net \cite{r82}, PoinTr \cite{r89} and other work \cite{r101, r102, r103}. As transformer outperforms various of methods in computer vision, the transformer-based methods will be widely studied in the next few years.
    \item Limited networks can fulfill robust real-time completion tasks. In addition, the network training process is time-consuming. Research emphasis should be focused on lightweight and compact structure design.
\end{itemize}

\section{Conclusion}
This paper has carried out a systematic review of the approaches for 3D point completion. Moreover, a comprehensive taxonomy and performance comparison of these approaches has been summed up. The advantages and limitations of each method are introduced, and the possible research directions are listed. This paper details DL's research challenges and opportunities in point cloud completion to promote the potential developments. Nowadays, nearly all methods are striving to tackle the two main challenges: fully exploit the structural information and predict fine-grained complete 3D shapes. Although significant progress has been made, many efforts still need to be paid to make point cloud completion practical.

\ifCLASSOPTIONcaptionsoff
  \newpage
\fi

\bibliographystyle{IEEEtran}
\bibliography{ref}

\vfill

\vfill

\end{document}